\documentclass[lettersize,journal]{IEEEtran}
\usepackage{amsmath,amsfonts}
\usepackage{algorithmic}
\usepackage{algorithm}
\usepackage{array}
\usepackage[caption=false,font=normalsize,labelfont=sf,textfont=sf]{subfig}
\usepackage{textcomp}
\usepackage{stfloats}
\usepackage{url}
\usepackage{soul}
\usepackage{verbatim}
\usepackage{graphicx}
\usepackage{cite}
\usepackage{booktabs} % for professional tables
\usepackage{adjustbox}

\usepackage{color}
\usepackage{comment}
\usepackage{amssymb} % for checkmark
\usepackage{pifont} % for xmark
\usepackage{multirow}
\usepackage{nicematrix}
\begin{document}

% \title{ Deep Transfer Learning in Turbofan Engine Prognostics: Methods, Evaluation and Future Directions}

% \title{A Survey on Deep Transfer Learning for Turbofan Engine Remaining Useful Life Prediction}

\title{Deep Domain Adaptation for Turbofan Engine Remaining Useful Life Prediction: Methodologies, Evaluation and Future Trends}

\author{Yucheng Wang*, Mohamed Ragab*, \IEEEmembership{Member, IEEE}, Yubo Hou, Zhenghua Chen, \IEEEmembership{Senior Member, IEEE}, \\
Min Wu, \IEEEmembership{Senior Member, IEEE}, and Xiaoli Li, \IEEEmembership{Fellow, IEEE} 
\IEEEcompsocitemizethanks{
    \IEEEcompsocthanksitem{* denotes equal contributions.}
        \IEEEcompsocthanksitem{Corresponding author: Min Wu (wumin\}@a-star.edu.sg).}
        \IEEEcompsocthanksitem{Yucheng Wang, Yubo Hou and Min Wu are with the Institute for Infocomm Research, Agency for Science, Technology and Research, Singapore (E-mail: \{wang\_yucheng, hou\_yubo, wumin\}@i2r.a-star.edu.sg).}
       \IEEEcompsocthanksitem{Mohamed Ragab is with Propulsion and Space Research Center, Technology Innovation Institute, UAE (E-mail: mohamedr002@e.ntu.edu.sg).} 
      \IEEEcompsocthanksitem{Zhenghua Chen is with the James Watt School of Engineering, University of Glasgow, UK (Email:Zhenghua.Chen@glasgow.ac.uk).} 
       \IEEEcompsocthanksitem{Xiaoli Li is with the ISTD Pillar at SUTD (E-mail: xiaoli\_li@sutd.edu.sg).} 
    }
}
    
\maketitle

\begin{abstract}
Remaining Useful Life (RUL) prediction for turbofan engines plays a vital role in predictive maintenance, ensuring operational safety and efficiency in aviation. Although data-driven approaches using machine learning and deep learning have shown potential, they face challenges such as limited data and distribution shifts caused by varying operating conditions. \textcolor{black}{Domain Adaptation (DA) has emerged as a promising solution, enabling knowledge transfer from source domains with abundant data to target domains with scarce data while mitigating distributional shifts.} Given the unique properties of turbofan engines—such as complex operating conditions, high-dimensional sensor data, and slower-changing signals—it is essential to conduct a focused review of DA techniques specifically tailored to turbofan engines. \textcolor{black}{To address this need, this paper provides a comprehensive review of DA solutions for turbofan engine RUL prediction, analyzing key methodologies, challenges, and recent advancements.} A novel taxonomy tailored to turbofan engines is introduced, organizing approaches into methodology-based (how DA is applied), alignment-based (where distributional shifts occur due to operational variations), and problem-based (why certain adaptations are needed to address specific challenges). This taxonomy offers a multidimensional view that goes beyond traditional classifications by accounting for the distinctive characteristics of turbofan engine data and \textcolor{black}{the standard process of applying DA techniques to this area}. Additionally, we evaluate selected DA techniques on turbofan engine datasets, \textcolor{black}{providing practical insights for practitioners and identifying key challenges.} Future research directions are identified to guide the development of more effective DA techniques, advancing the state of RUL prediction for turbofan engines.
%The source codes and experimental data used in this paper are available at: \url{https://github.com/keyplay/DA_RUL_Baseline}. 

\end{abstract}

\section{Introduction}
%%%% Narrative %%%%
% Introduction to turbofan engines 
% Why we focus on turbofan engines
% Data-driven approaches for turbofan engines 
% Limitation of data-driven approaches
% UDA for turbofan engines 
% Existing works for DA for RUL for turbofan engines. 
% Mention that here we aim to propose a review paper
% Contrasting with existing review papers 
% Introduction of our review paper and its content 

% Motivation and Importance:
\textcolor{black}{Turbofan engines serve as the backbone of modern aviation, powering aircraft and requiring peak performance~\cite{zhao2017remaining,hanachi2018performance,li2023remaining}.} However, nearly 60\% of aircraft breakdowns can be attributed to turbofan engine failures, \textcolor{black}{underscoring the urgent need for advanced maintenance strategies that go beyond traditional corrective or preventive approaches~\cite{tuzcu2021energy}.} Prognostics and Health Management (PHM) concepts, including \textcolor{black}{Condition-Based Maintenance (CBM) and Predictive Maintenance (PM)}, \textcolor{black}{provide a transformative approach to ensuring operational integrity, reducing maintenance costs, and extending engine life~\cite{hu2022prognostics,zonta2020predictive,rath2024aero}.}

\textcolor{black}{The prediction of a turbofan engine's Remaining Useful Life (RUL) is a cornerstone of predictive maintenance, enabling data-driven decisions that enhance operational safety and efficiency~\cite{mao2022self,mitici2023dynamic,chen2021risk,lin2021novel,jiao2021remaining,li2024sensor,piao2024crulp}.} This task has traditionally been approached through physical and statistical model-based methods, as well as data-driven techniques. \textcolor{black}{While physical models offer detailed insights, they often struggle with the complexity and noise inherent in turbofan engine operations. Similarly, statistical models, though useful, frequently depend on expert knowledge and may not fully leverage the wealth of available sensor data~\cite{lei2018machinery,brunton2021data}.}

% data driven approaches 
\textcolor{black}{Data-driven methods, empowered by advancements in machine learning and deep learning, have shown significant promise in predicting RUL by leveraging sensor data without requiring an in-depth understanding of the underlying degradation mechanisms~\cite{zhao2019deep,de2023data,liu2022aircraft,kordestani2023overview,jiao2021remaining,li2022remaining39}.} However, the effectiveness of these approaches is often constrained by the need for extensive labeled datasets. \textcolor{black}{Due to safety and cost considerations, turbofan engines are rarely operated until complete failure, making it difficult to collect sufficient labeled data for model training.} Furthermore, turbofan engines operate under diverse conditions, such as variations in flight phases, which lead to shifts in data distribution. \textcolor{black}{Adapting models to these distributional shifts would require collecting and labeling large datasets for each new operating condition, a process that is both impractical and resource-intensive.} These challenges \textcolor{black}{pose significant limitations on the applicability of conventional data-driven methods for RUL prediction in turbofan engines~\cite{ragab2020contrastive}.}

% Domain adaptation for RUL
\textcolor{black}{Domain Adaptation (DA) offers a promising solution to these challenges by enabling knowledge transfer from a source domain to a target domain, effectively addressing distribution shifts caused by varying operating conditions.} By leveraging DA techniques, a model can be trained on a dataset with abundant labeled samples, such as simulated datasets, and then adapted to the target domain. \textcolor{black}{This strategy significantly diminishes the need for extensive labeling in the target domain.} While DA has achieved remarkable success in computer vision applications where labeled data is scarce \cite{zhao2020review,liu2022deep,wang2018deep,chen2024dynamic}, \textcolor{black}{it is gaining traction in PHM for turbofan engines, improving RUL prediction under diverse operating conditions \cite{ding2022transfer,fan2020transfer,zhang2018transfer,gribbestad2021transfer,yang2024label,ma2023estimating}.} Despite these advancements, current research in DA for RUL prediction remains fragmented, \textcolor{black}{characterized by inconsistent problem definitions and limited cohesion across studies.}

% comparison with existing reviews
\textcolor{black}{Currently, numerous reviews have been published on RUL prediction, primarily focusing on conventional methods} such as physics-based models \cite{cubillo2016review}, traditional machine learning techniques \cite{zhang2018degradation, kan2015review}, and deep learning architectures \cite{lei2018machinery, zhao2019deep}. However, these surveys often overlook the specific requirements of DA for turbofan engines, \textcolor{black}{failing to address critical challenges such as changing operating conditions and limited data availability.} While some recent reviews explore DA techniques \cite{dong2021transfer, chen2023transfer, yao2023survey}, \textcolor{black}{they predominantly focus on small components like bearings and gears, leaving a gap in the comprehensive analysis of DA methods specifically tailored for turbofan engine RUL prediction.} Furthermore, these reviews do not introduce new taxonomies to categorize DA approaches \textcolor{black}{based on the unique operational conditions and data characteristics of turbofan engines.}

% The challenges in turbofan engines
\textcolor{black}{Turbofan engines differ from gears and bearings in three key aspects.} First, they operate under more complex conditions, including variations in altitude, temperature, and flight phase, with the same engine encountering diverse scenarios throughout its lifetime. This amplifies the domain shift between operating conditions, making adaptation more challenging. Second, data from turbofan engines is high-dimensional, with more sensors required to monitor engine status compared to bearings or batteries. The large number of channels results in a complex feature space, complicating the modeling of dependencies and domain alignment. Third, unlike bearings, which generate high-frequency signals such as vibrations or acoustic data that demand high temporal resolution, turbofan engines produce slower-changing signals (e.g., temperature, pressure, fuel flow) that reflect overall health rather than immediate conditions, necessitating a lower subsampling rate. \textcolor{black}{Given these distinct characteristics, a focused survey of DA techniques for turbofan engine RUL prediction is essential. Such a survey would not only address the unique challenges of domain shifts and data limitations in engine operation but also identify effective strategies for adaptation.} This effort would provide clearer insights into the development of DA frameworks that are both theoretically rigorous and practically applicable in aviation, \textcolor{black}{guiding future research toward solutions tailored to the industry's specific needs.}

% proposed approach 
This paper presents a focused review of DA for RUL prediction in turbofan engines, examining key methodologies, challenges, and recent advancements. Our literature analysis clarifies the current state of DA in this critical domain and identifies future opportunities to enhance both reliability and safety in aviation. \textcolor{black}{To provide a comprehensive review, we propose a novel taxonomy tailored to turbofan engines,} categorizing existing works into methodology-based approaches (how DA is performed), alignment-based approaches (where distributional shifts occur due to varying operational conditions), and problem-based approaches (why specific adaptations are required to address unique challenges). \textcolor{black}{This framework extends beyond traditional classifications by aligning with the standard process of applying DA to turbofan engine RUL prediction} and accounting for the distinct characteristics of turbofan engine data, offering a structured and multidimensional perspective on DA techniques. Additionally, we benchmark selected DA techniques for turbofan engines, providing a resource for newcomers to navigate this research area effectively. We also analyze the limitations of current DA methods in real-world applications and propose solutions to improve their reliability and effectiveness for turbofan engine RUL prediction. This comprehensive review aims to guide future research toward developing more robust and practical DA frameworks for the complexities of aviation prognostics.

\begin{figure*}
    \centering

    \includegraphics[width = 1\linewidth]{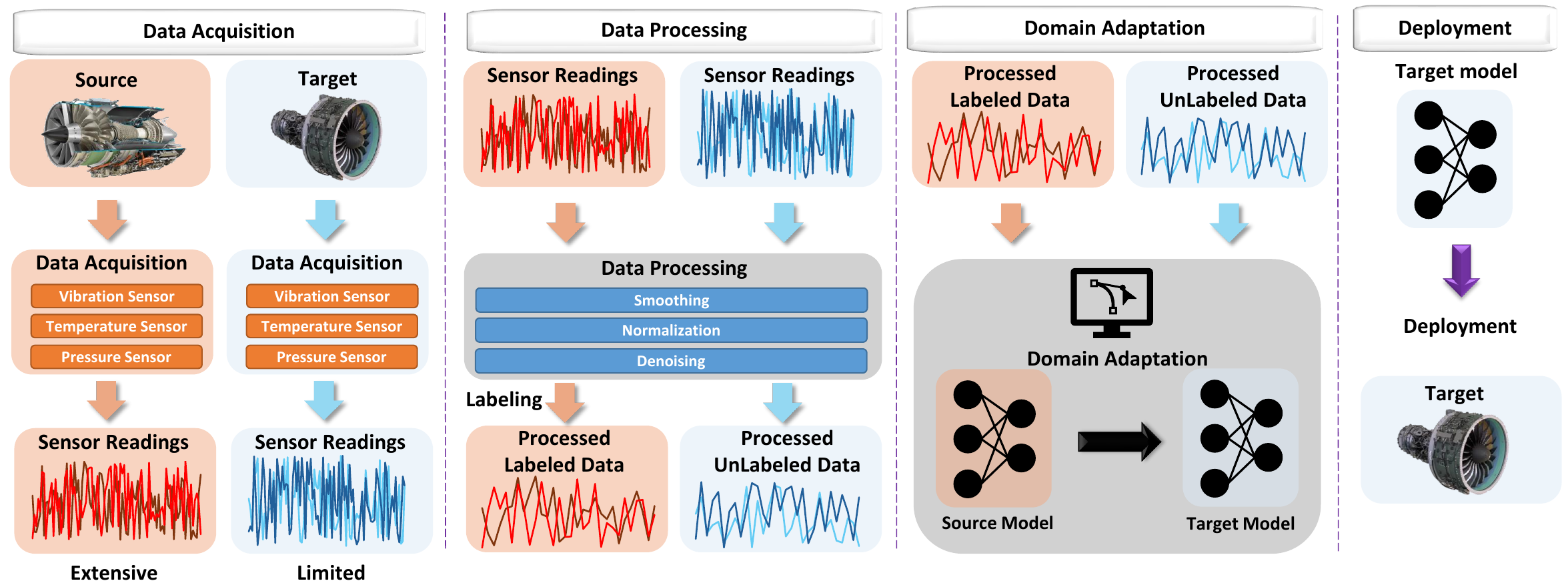}
    \caption{Overall workflow of DA for RUL prediction. \textcolor{black}{Data is first collected in the source and target domains. After data processing, DA techniques have been adopted to reduce the gaps between domains. After adaptation, the adapted model can then be used in the target domain for RUL prediction.}
}
        \label{fig:overallflow}

\end{figure*}

The key contributions are summarized as follows:
\begin{itemize}
    \item \textbf{Focused Review:} Presenting comprehensive analysis of DA specifically for turbofan engine RUL prediction, highlighting recent advancements and current best practices.
    \item \textbf{Novel Taxonomy:} Introducing a new taxonomy tailored to the unique aspects of turbofan engines, enhancing analytical depth in DA applications.
    \item \textbf{Open-Source Collection:} Providing an evaluation for DA techniques on turbofan engine data, facilitating research engagement and practical testing.
    \item \textbf{Research Opportunities:} Identifing future research directions to improve aviation safety and reliability through enhanced predictive maintenance strategies.
\end{itemize}

%% Difference to Existing Reviews

% \begin{table}[]
%     \centering
%     \caption{Different problem settings of DA in RUL prediction}
%     \resizebox{0.5\textwidth}{!}{
%     \begin{tabular}{@{}lccc@{}}
%     \toprule
%     Scenario                           & Source Data              & Target Data                         & Distribution Shift \\ 
%     \midrule
%     Deep Learning                      & Fully labeled            & Labeled                         & \xmark             \\
%     Supervised Domain Adaptation       & Fully labeled            & Labeled                             & \cmark             \\
%     Semi-supervised Domain Adaptation  & Fully labeled            & Partially labeled \& Unlabeled      & \cmark             \\
%     Unsupervised Domain Adaptation     & Fully labeled            & Unlabeled                           & \cmark             \\
%     % Domain Generalization              & Fully labeled            & Unavailable                         & \cmark             \\ 
%     \bottomrule
%     \end{tabular}%
%     }
%     \label{tab:problem_setting}
% \end{table}

\section{Preliminaries}

When exploring DA for RUL prediction in turbofan engines, it is essential to establish a foundational understanding of the key concepts and terminologies that underpin this focused topic. This section provides an overview of the preliminaries, focusing on turbofan engines and key terminologies in DA.
\begin{itemize}
    \item \textit{Turbofan Engine}: \textcolor{black}{Turbofan engines, widely used in jet aircraft, generate thrust by directing airflow through a core engine and a bypass fan.} The Low-Pressure Compressor (LPC) and High-Pressure Compressor (HPC) sequentially compress air for combustion. The resulting high-energy exhaust gases drive the High-Pressure Turbine (HPT) to power the HPC and the Low-Pressure Turbine (LPT) to power the fan and LPC. \textcolor{black}{Most of the airflow bypasses the core through the bypass duct, which enhances thrust and reduces noise.}

    \item \textit{Sensor}: \textcolor{black}{Due to their complexity} , turbofan engines are equipped with multiple sensors to monitor key parameters, ensuring enhanced performance and early fault detection.These sensors collect real-time data essential for engine control and maintenance. Common types include temperature, pressure, and speed sensors, \textcolor{black}{which collectively provide a comprehensive view of the engine’s condition.}

    \item \textit{Domain}: In the realm of DA, a domain $\mathcal{D}$ is characterized by a feature space $\mathcal{X}$ and a marginal probability distribution $P(X)$, where $X = \{x_1, \dots, x_n\} \in \mathcal{X}$. The feature space represents all possible inputs, while $P(X)$ denotes the distribution of these inputs within the space.
    
    \item \textit{Task}: A task $\mathcal{T}$, in this context, is defined by a label space $\mathcal{Y}$ and a predictive function $f(\cdot)$ that maps inputs to labels, denoted as $f(\cdot): \mathcal{X} \rightarrow \mathcal{Y}$.
    
    \item \textit{Deep Learning}: Deep learning in DA involves a training domain $\mathcal{D}_{\text{train}}$ and a task $\mathcal{T}_{\text{train}}$. The goal is to use the training data to learn a function $f(\cdot)$ that can be effectively applied to test data from a domain $\mathcal{D}_{\text{test}}$ and task $\mathcal{T}_{\text{test}}$, with an emphasis on domains and tasks that are either similar or identical between training and testing phases.
        
    \item \textit{Domain Adaptation}: DA involves a source domain $\mathcal{D}_S$ with its associated task $\mathcal{T}_S$, and a target domain $\mathcal{D}_T$ with task $\mathcal{T}_T$. Given the presence of domain discrepancies, i.e., $\mathcal{D}_S \neq \mathcal{D}_T$, the primary objective is to transfer knowledge from the source domain to the target domain. This process can encompass scenarios where the source domain contains labeled samples, while the target domain may have either limited labeled samples (supervised DA) or unlabeled samples (unsupervised DA).
    
    % \item \textit{Domain Generalization}: Another subset of TL, domain generalization, involves training on multiple source domains $\{\mathcal{D}_S\}_{i=1}^{M}$, with the aim of generalizing well to an unseen target domain $\mathcal{D}_T$.
\end{itemize}

These concepts form the backbone of our exploration into DA for RUL prediction, providing the necessary framework to understand its applications and implications in the domain of turbofan engine maintenance.

\begin{figure*}[htbp!]
    \centering
    \includegraphics[width = 0.95\linewidth]{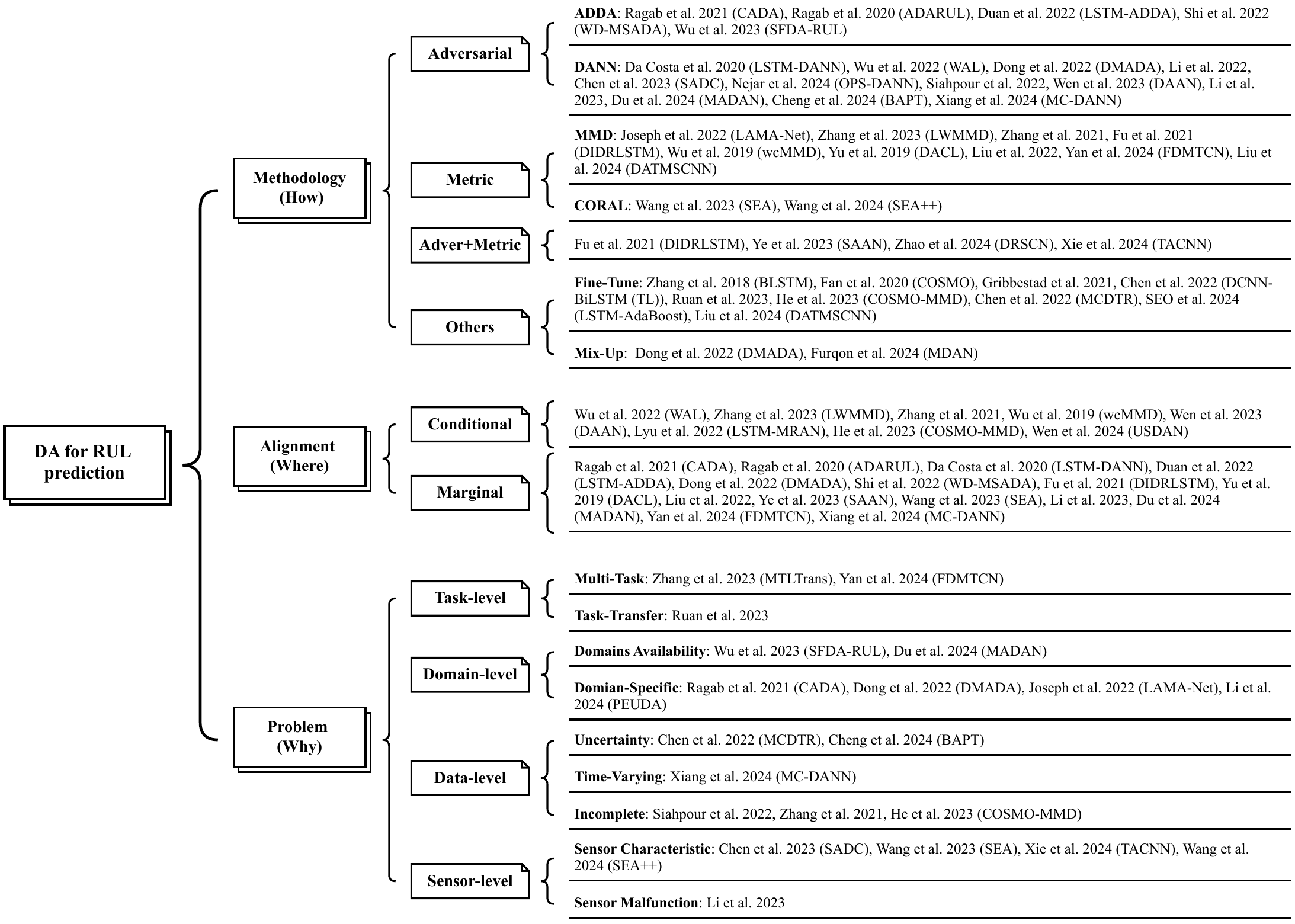}
    \caption{Overall summary of DA methods for turbofan engine RUL prediction.
}
    \label{fig:overall}
\end{figure*}

\section{Domain Adaptation for RUL prediction}

\subsection{Workflow}
The overall workflow of DA for turbofan engine RUL prediction is illustrated in Fig. \ref{fig:overallflow}. Initially, we possess ample data acquired from the source domain, while the number of samples from the target domain is limited. Given that the limited target domain samples may impede the performance of a deep learning model, it becomes imperative to employ DA models to transfer knowledge from the source domain to the target domain.
Following data preprocessing in both domains, DA models are applied. These models leverage the knowledge contained within the source domain data to enhance the target model's capacity to learn effective features. By effectively aligning the feature distributions between domains and mitigating the effects of domain shift, DA facilitates the adaptation of the target model to the characteristics of the target domain.
Subsequently, the improved target model, enriched with knowledge from the source domain, can make accurate predictions of remaining useful life in practical applications.

% \vspace{-0.2cm}
\subsection{Taxonomy}
% This paper introduces a new taxonomy of domain adaptation approaches tailored for Remaining Useful Life (RUL) prediction. The taxonomy categorizes these approaches into three principal groups based on the how, where, and what of the alignment process. Methodology-based Approaches focus on the technical and procedural methods employed to align different domains, emphasizing the algorithmic strategies that facilitate domain adaptation. Alignment-based Approaches delve into deciding where to apply the alignment methodology, highlighting the specific or general conditions under which domain alignment is most effective, thus influencing the applicability and success of the adaptation. Lastly, Problem-based Approaches are designed to address the unique challenges and characteristics inherent to RUL prediction, focusing on adapting to the specific problems presented by the application environment. By categorizing domain adaptation approaches into these detailed groups, this taxonomy aims to enhance the precision and applicability of domain adaptation techniques in the diverse field of RUL prediction, providing a structured framework that details not only how to align domains but also where and what aspects should be aligned.
\textcolor{black}{To comprehensively summarize existing works in this area, it is essential to analyze the standard process of applying DA to turbofan engine RUL prediction, which consists of three key steps: 1. Selecting suitable DA techniques (How): The first step of this process is to choose an appropriate DA methodology, such as adversarial learning or metric-based approaches, to address distribution shifts. 2. Identifying where adaptation is needed (Where): Once a DA method is selected, it is crucial to determine where it should be applied—whether these shifts occur in feature distributions (marginal alignment) or conditionally on labels (conditional alignment). 3. Addressing unique challenges in turbofan engine RUL prediction (Why): Unlike general DA applications, turbofan engine RUL prediction presents unique challenges, such as the problems in task, domain, data, and sensor-level. Thus, problem-driven adaptations are necessary.}

\textcolor{black}{Aligning with the fundamental steps in the standard process, we propose a novel taxonomy that categorizes existing works into methodology-based, alignment-based, and problem-based methods, aiming to provide a comprehensive framework for summarizing existing works in this area while ensuring they remain natural and complementary. Here,} methodology-based methods focus on \textit{how} DA is performed for turbofan engines; alignment-based methods focus on addressing \textit{where} distributional shifts occur due to varying operational conditions; and problem-based methods focus on explaining \textit{why} certain adaptations are necessary based on unique challenges. This classification goes beyond traditional approaches \textcolor{black}{by considering the process of applying DA into turbofan engine RUL prediction} and recognizing the distinctive characteristics of turbofan engine data and applications, offering a more comprehensive and multidimensional view of DA techniques.

\begin{itemize}
  \item \textbf{Methodology-based Approaches (How):} Focus on the technical methods used to align different domains, highlighting the algorithmic strategies that drive effective DA. By categorizing approaches into adversarial, metric, adversarial+metric, and other methods that do not fit these groups, this classification clarifies the strengths and limitations of each with respect to the unique demands of turbofan engine RUL prediction.

\item \textbf{Alignment-based Approaches (Where):} Identify the appropriate contexts for applying alignment methods, focusing on the conditions under which domain alignment is most effective. By distinguishing between marginal and conditional alignment, this classification captures which data shifts are addressed—whether aligning global feature distributions (marginal) or ensuring consistent behavior across label-conditioned groups (conditional). \textcolor{black}{Understanding where DA is required enables researchers and practitioners to fine-tune their strategies, enhancing the effectiveness and applicability of adaptation efforts.}

\item \textbf{Problem-based Approaches (Why):} Address the unique challenges of RUL prediction for turbofan engines, focusing on adapting to the specific problems posed by the application environment. By dividing approaches into four levels—Task-level, Domain-level, Data-level, and Sensor-level—this framework captures both high-level and low-level challenges. \textcolor{black}{This problem-centric approach ensures that DA solutions are not only theoretically robust but also practically applicable in the complex context of turbofan engine maintenance and operations.}
  \end{itemize}
% By categorizing domain adaptation approaches into these detailed groups, this taxonomy aims to enhance the precision and applicability of domain adaptation techniques in the diverse field of RUL prediction, providing a structured framework that details not only how to align domains but also where and what aspects should be aligned.

% In recent years, dozens of methods have been proposed to address the domain shift problem for RUL prediction of turbofan engines, while considering the label scarcity. In our work, we introduce a new taxonomy for existing works as shown in Fig. \ref{fig:overall}, including the used methodology, the alignment strategy, and the RUL-problem specific approaches.

\subsection{Methodology-based approaches}
We begin by introducing methodology-based approaches, \textcolor{black}{which can be broadly categorized into adversarial, metric-based, hybrid adversarial-metric, and other methods that do not fit within these primary categories, as illustrated in Fig. \ref{fig:overall}. The core principles of these approaches, along with their application in existing DA works, are discussed in detail in the following sections.}

\subsubsection{Adversarial}
The adversarial-based methods have emerged as one of the most effective approaches in the realm of DA for \textcolor{black}{turbofan engine RUL prediction}, consistently demonstrating superior performance. The main idea of these methods is to train an encoder and a domain classifier concurrently. The domain classifier discerns the origin domain of a sample, while the encoder endeavors to deceive the domain classifier by learning domain-invariant representations. The existing adversarial-based methods mainly include two categories, Domain-Adversarial Neural Networks (DANN) \cite{Ganin2017Domain} and Adversarial Discriminative Domain Adaptation (ADDA) \cite{tzeng2017adversarial} and their variants \cite{long2018conditional,yu2019transfer,gao2021gradient,chen2022reusing}.

\begin{figure}[b]
    \centering
    \includegraphics[width = 1.\linewidth]{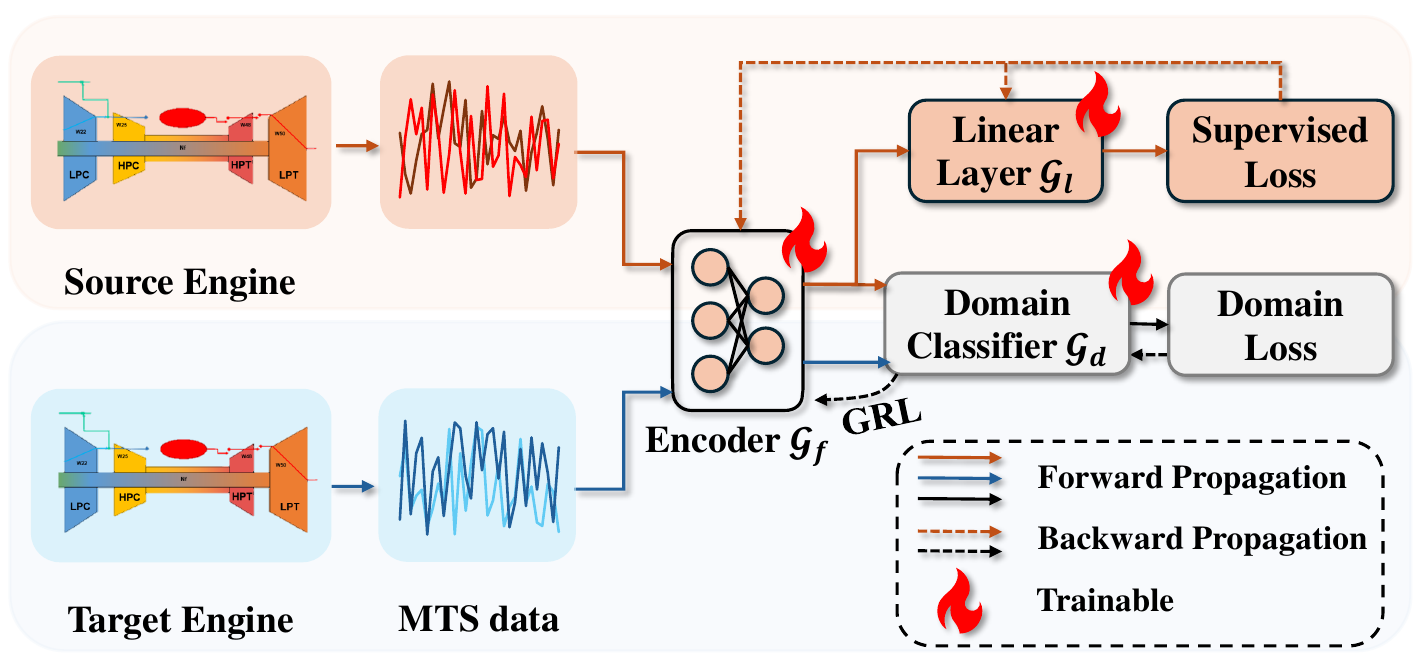}
    \caption{\textcolor{black}{Workflow of DANN for RUL prediction. The encoder extracts features from both source and target domains, while the domain discriminator tries to distinguish them. A Gradient Reversal Layer (GRL) ensures the encoder learns domain-invariant features by reversing the discriminator’s gradients. Simultaneously, a task-specific linear layer minimizes prediction error on the source domain to improve generalization to the target domain.}}
    \label{fig:DANN}
\end{figure}

\paragraph{DANN}
Fig. \ref{fig:DANN} illustrates the architecture of DANN, comprising an encoder $\mathcal{G}_f(x_i;\theta_f)$, a linear layer $\mathcal{G}_l(x_i;\theta_l)$, and a domain classifier $\mathcal{G}_d(x_i;\theta_d)$, where $\theta_l$, $\theta_f$, and $\theta_d$ are trainable parameters. The linear layer is responsible for classifying features acquired by the encoder. Concurrently, the domain classifier $\mathcal{G}_d(x_i;\theta_d)$ categorizes features from of samples from which domain. Consequently, DANN incorporates two loss functions: one for supervised learning in the source domain and another for DA, as depicted as follows.
\begin{equation}
    \label{eq:dann}
    \begin{split}
     \mathcal{L} &= \mathcal{L}_{S} + \mathcal{L}_{D},\\
    \mathcal{L}_S &= \sum_{i\in\mathcal{D}_S}MSE(\mathcal{G}_l(\mathcal{G}_f(x_i;\theta_f);\theta_l)),\\
        \mathcal{L}_D &= -\sum_{i\in\{\mathcal{D}_S, \mathcal{D}_T\}}CE(\mathcal{G}_d(\mathcal{G}_f(x_i;\theta_f);\theta_d).
    \end{split}
\end{equation}
In Eq. (\ref{eq:dann}), the supervised loss in the source domain endeavors to minimize the discrepancy between predicted and actual sample labels. Conversely, the domain loss aims to maximize the disparity between predicted domain labels and actual domain labels, thereby encouraging the learning of features that can deceive the domain classifier. Simultaneously, the domain classifier aims to minimize the domain loss by improving its ability to correctly classify domain labels. Based on the objectives, the optimization goal can be formally represented as the Eq. (\ref{eq:dann_op}), where the parameters $\theta_f$ and $\theta_l$ of the encoder and the linear layer are trained to minimize the final loss, while the parameter $\theta_d$ of the domain classifier is trained to maxmize the final loss. 
\begin{equation}
    \label{eq:dann_op}
    \begin{split}
        (\hat{\theta}_f, \hat{\theta}_l) &= argmin_{({\theta}_f,{\theta}_l)}\mathcal{L},\\
        \hat{\theta}_d &= argmax_{({\theta}_d)}\mathcal{L}.
    \end{split}
\end{equation}

From Eq. (\ref{eq:dann}) and (\ref{eq:dann_op}), we observe that the domain classifier strives to minimize the domain loss, while simultaneously, the encoder aims to maximize it, making the adversarial nature of the gradients of the encoder and domain classifier. To facilitate end-to-end training, DANN introduces the Gradient Reversal Layer (GRL), which effectively reverses the gradient flow when propagating the domain loss back to the encoder. This mechanism enables seamless integration of DA into the training process, ensuring the encoder learns domain-invariant representations effectively.

\textcolor{black}{Currently, DANN has gained widespread adoption in this domain \cite{wu2022weighted,dong2022dual,li2022domain,nejjar2024domain}.} The DANN-based framework proposed by Da Costa et al. \cite{da2020remaining} pioneered this field, being the first to utilize LSTM as the base encoder and DANN to mitigate domain differences for turbofan engine RUL prediction. Since then, a diverse array of subsequent studies have been inspired \cite{dong2022dual,li2022domain,joseph2022lama,chen2023aero,zhang2023variational,siahpour2022novel,zhang2021transfer}. DANN was further advanced by Wu et al. \cite{wu2022weighted} through the introduction of a weighted adversarial loss. This innovation involves assigning small weights when significant differences exist between the RULs of the samples in two domains. Notably, in the absence of labels in the target domain, pseudo-labels are generated using encoders trained on the source domain. \textcolor{black}{Additionally, DANN was enhanced by Dong et al. \cite{dong2022dual} through the introduction of mix-up operations, specifically domain mix-up.} This technique generates pseudo-samples from each domain and then employs adversarial-based methods to ensure the encoder cannot distinguish which domain a mixed sample originates from. The work proposed by Li et al. \cite{li2022domain} argued that the entire life cycle of machines should follow a similar path—health, initial degradation, and severe degradation. Thus, they introduced a semantic-level alignment based on DANN to adapt to the underlying structure of the machine life cycle. \textcolor{black}{Furthermore, the work by Nejjar et al. \cite{nejjar2024domain} highlighted that an entire flight comprises distinct phases, such as take-off, cruise, and descent.} Recognizing that each phase possesses different characteristics leading to diverse marginal distributions, the authors argued for treating each phase as a sub-domain and aligning them individually. To achieve this, they employed DANN, utilizing multiple discriminators to discern the origin of samples, indicating from which specific phase each sample originated.

These DANN-based methods have demonstrated the effectiveness of adversarial approaches in reducing domain discrepancies for turbofan engine RUL prediction. Moreover, their end-to-end training framework makes them easy to implement. As shown in Fig. \ref{fig:overall}, DANN-based methods are notably more popular than ADDA. However, they come with a key limitation: balancing the two losses—one for source domain prediction and one for DA, as shown in Eq. (\ref{eq:dann})—can be challenging. This trade-off may hinder optimal adaptation and prevent the encoder from learning the most suitable features for the target domain, \textcolor{black}{hindering their applications in real-world turbofan engine applications.}
\begin{figure}[b]
    \centering
    \includegraphics[width = 1.\linewidth]{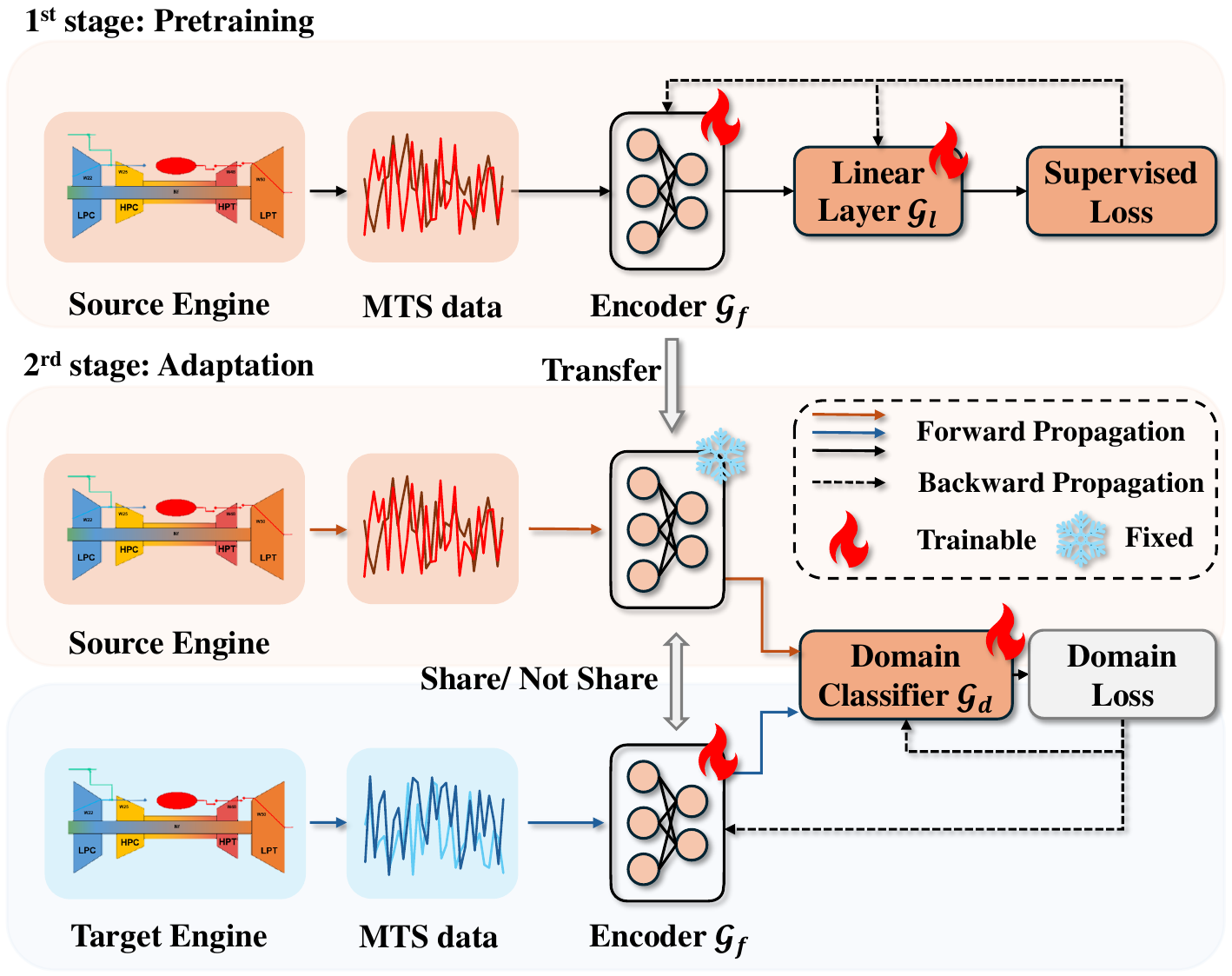}
    \caption{\textcolor{black}{Workflow of ADDA for RUL prediction. A two-stage adversarial framework where the first stage focuses on pretraining the encoder on labeled source data. In the second stage, the pretrained encoder is adapted to the target domain through adversarial learning. The encoder is fine-tuned to produce domain-invariant features, while the domain discriminator is trained to differentiate between the two domains.}}
    \label{fig:ADDA}
\end{figure}
\paragraph{ADDA} \textcolor{black}{As a two-stage adversarial approach, ADDA can address the challenge of balancing source training and adaptation by dividing the process into two distinct stages: pretraining and adaptation. This separation allows the model to focus more effectively on adapting a pretrained encoder to learn domain-invariant features.} Illustrated in Fig. \ref{fig:ADDA}, the initial stage involves pretraining the model with labeled samples from the source domain. Subsequently, the pretrained encoder is transferred to the target domain. In the adaptation stage, the encoder is fine-tuned to deceive the domain classifier, while simultaneously, the domain classifier is trained to discern the domain of each sample accurately. The loss functions are shown in Eq. (\ref{eq:adda}).
\begin{equation}
    \label{eq:adda}
    \begin{split}
        \mathcal{L}_{D} = 
        &-\sum_{i\in\mathcal{D_S}}log\mathcal{G}_d(\mathcal{G}_f(x_i;\theta_f);\theta_d)\\
        &-\sum_{i\in\mathcal{D_T}}log(1-\mathcal{G}_d(\mathcal{G}_f(x_i;\theta_f);\theta_d)),\\
        \mathcal{L}_{M} = 
        &-\sum_{i\in\mathcal{D_T}}log\mathcal{G}_d(\mathcal{G}_f(x_i;\theta_f);\theta_d).\\
    \end{split}
\end{equation}
where the parameters $\theta_f$ and $\theta_d$ are trainable for the encoder and domain classifier respectively. It is noted that the linear layer has been trained in the pretraining stage, and it is fixed in the adaptation stage. 

% \begin{equation}
%     \label{eq:adda}
%     \begin{split}
%         min_{\theta_d}\mathcal{L}_{D} = \\
%         &-\sum_{i\in\mathcal{D_S}}log\mathcal{G}_d(\mathcal{G}_f(x_i;\theta_f);\theta_d)\\
%         &-\sum_{i\in\mathcal{D_T}}log(1-\mathcal{G}_d(\mathcal{G}_f(x_i;\theta_f);\theta_d))\\
%         min_{\theta_f}\mathcal{L}_{M} = \\
%         &-\sum_{i\in\mathcal{D_T}}log\mathcal{G}_d(\mathcal{G}_f(x_i;\theta_f);\theta_d)\\
%     \end{split}
% \end{equation}

\textcolor{black}{Currently, several works \cite{ragab2020contrastive,ragab2020adversarial,wu2023privacy,shi2023wasserstein,lv2020sequence} focus on adapting ADDA for aerospace RUL prediction.} For example, an ADDA-based framework was proposed by Ragab et al. \cite{ragab2020adversarial} to adapt a LSTM encoder to target domain samples, where the LSTM encoder was pretrained with labeled RUL samples in a source domain. Then, the Wasserstein distance was introduced by Shi et al. \cite{shi2023wasserstein} as a metric in their approach. Through adversarial-based optimization, they aimed to optimize the parameters of encoders and discriminators to minimize the Wasserstein distance between domains. \textcolor{black}{Moreover, the concept of ADDA—using adversarial fine-tuning—has been extended in an ADDA-like framework by Wu et al. \cite{wu2023privacy}. In their approach, a pretrained source encoder is leveraged to fine-tune both the encoder and two predictors in the target domain through a dual-objective strategy. When the encoder is fixed, the predictors are adjusted to maximize their dissimilarity. Conversely, with the predictors fixed, the encoder is fine-tuned to minimize the dissimilarity between them. This iterative process aims to identify target domain samples that are far from the source domain’s support, enhancing the adaptation to the target domain. In addition to being applied directly for DA, ADDA has been used to generate datasets from auxiliary sources, as demonstrated by Lv et al. \cite{lv2020sequence}, where the generated data is utilized for cross-domain scenarios. Beyond aircraft engines, adversarial-based methods have also found extensive application in bearing RUL prediction \cite{hu2022remaining,wang2021bearing,zhuang2022adversarial,li2020data,zou2022method}, showcasing their versatility in different predictive maintenance tasks.}

Compared to other DA methods, adversarial-based approaches are more effective at transferring highly nonlinear information by leveraging a discriminator network to ensure that features from the source and target domains are indistinguishable. This capability allows to handle complex, nonlinear distribution shifts through flexible, sophisticated mappings. Thus, these methods are well-suited for capturing intricate relationships in turbofan engine data across varying operational regimes. They are particularly advantageous when source and target domains differ significantly in operational conditions and failure modes, as they can dynamically adapt to such complexities. 

\textcolor{black}{However, these methods face several limitations that can significantly impact RUL prediction in turbofan engines. First, they are computationally intensive, requiring the simultaneous training of both a feature extractor and a discriminator, which increases training time and demands careful hyperparameter tuning. In real-world applications where timely model deployment is crucial for maintenance decision-making, this computational overhead can hinder practical implementation. Second, adversarial-based approaches are prone to mode collapse \cite{kossale2022mode}, where the model learns to align only a subset of features while ignoring essential degradation patterns. This is particularly problematic for RUL prediction, as ignoring critical health indicators (e.g., temperature, pressure, vibration trends) may lead to erroneous predictions, reducing reliability in predictive maintenance strategies. Moreover, instability in adversarial training can lead to poor convergence, making it difficult to ensure consistent performance across different turbofan engine datasets. These challenges highlight the need for improved stability mechanisms, such as hybrid approaches, to make adversarial DA more reliable and practical for real-world turbofan engine health monitoring.}

\subsubsection{Metric-based}
\begin{figure}[t]
    \centering
    \includegraphics[width = 1.\linewidth]{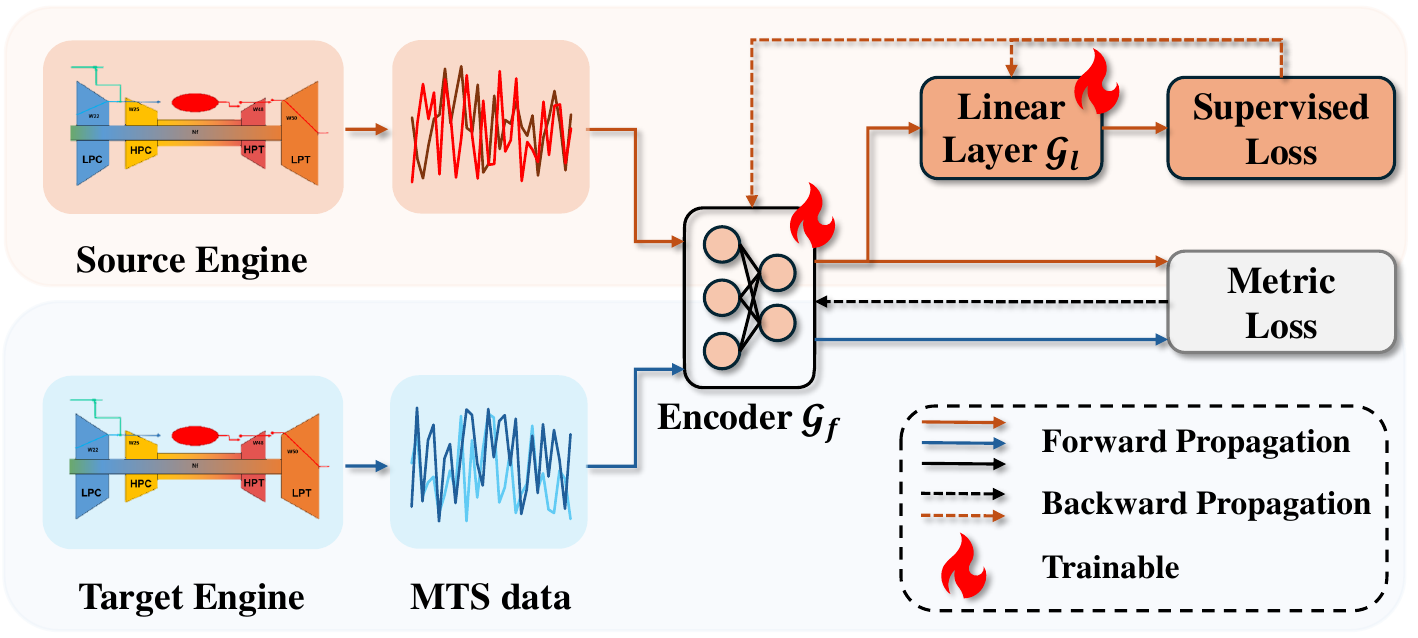}
    \caption{\textcolor{black}{Workflow of metric-based methods for RUL prediction. Metric loss is minimized for reducing domain discrepancies.}}
    \label{fig:metric}
\end{figure}

\textcolor{black}{Metric-based methods effectively address the challenges of adversarial-based approaches.} By directly minimizing the distance between source and target feature distributions, they ensure overall alignment across domains and reduce the risk of mode collapse. Furthermore, these methods are simpler to train, as they rely on predefined metrics rather than complex adversarial mechanisms. \textcolor{black}{This simplicity often leads to faster convergence and lower computational overhead, making them a more efficient solution for DA.} The general metric-based method is illustrated in Fig. \ref{fig:metric}, where the metric loss is measured by distributional distances. Currently, prominent metric-based approaches include Maximum Mean Discrepancy (MMD) \cite{tzeng2014deep} and Deep CORrelation ALignment (CORAL) \cite{sun2017correlation}, as well as their variants \cite{long2017deep,long2018transferable,long2015learning,rozantsev2018beyond}.

\paragraph{MMD}
MMD is based on comparing the means of representations of two domains. It quantifies the dissimilarity between the distributions of two datasets by computing the difference with features transformed into a reproducing kernel Hilbert space (RKHS). Given two datasets $\mathcal{D}_S$ and $\mathcal{D}_T$ drawn from distributions $\mathcal{P}_S$ and $\mathcal{P}_T$ respectively, and a feature map $\phi:\mathcal{D}\to\mathcal{H}$ that maps samples into a reproducing kernel Hilbert space $\mathcal{H}$, MMD can be defined defined as:
\begin{equation}
    \label{eq:mmd}
    \text{MMD}(\mathcal{P}_S, \mathcal{P}_T) = \sup_{\|f\|_{\mathcal{H}} \leq 1} (\mathbb{E}_{x_S \sim \mathcal{P}_S}[f(x_S)] - \mathbb{E}_{x_T \sim \mathcal{P}_T}[f(x_T)])
\end{equation}
where $f$ is a function in the RKHS $\mathcal{H}$ with a norm less than or equal to 1. This formula essentially calculates the maximum of the difference in the means of the function $f$ evaluated at samples drawn from distributions $\mathcal{P}_S$ and $\mathcal{P}_T$. In practice, MMD is often computed using a kernel function $k(x,y)$ to measure the similarity between samples $x$ and $y$ in the feature space. Common choices for the kernel function include Gaussian kernel, polynomial kernel, and linear kernel.

In recent years, MMD has been utilized for reducing domain discrepancy \cite{joseph2022lama,zhang2023variational,zhang2021transfer,wu2019weighted,yu2019domain,liu2022cnn}. The method was firstly introduced by Yu et al. \cite{yu2019domain} for turbofan engine RUL prediction, employing an LSTM-CNN-based framework as the encoder. Following by this, a CNN-based framework with MMD by Liu et al. \cite{liu2022cnn} to aligned source and target domains. An encoder-decoder framework with MMD was then designed by Joseph and Lalwani \cite{joseph2022lama} for alignment of encoder and decoder outputs. Next, the vanilla MMD has been improved to achieve better alignment. For example, the working condition-wise alignment has been introduced by Wu et al. \cite{wu2019weighted} to improve MMD, namely wcMMD, to align each working condition. A weighted MMD was then introduced by Zhang et al. \cite{zhang2023variational} to align segmented machine life cycles. Furthermore, MMD has also been applied to address incomplete RUL. A degradation fusion module was designed by Zheng et al. \cite{zhang2021transfer}, aiming to align the source and target degradation state data by using the MMD metric.

\textcolor{black}{With a a solid theoretical basis as a measure of the difference between two distributions in a RKHS, MMD provides a quantitative, interpretable measure of the discrepancy between distributions, helping to track domain alignment during training. However, the performance of MMD is sensitive to the choice of kernel and hyperparameters, requiring careful tuning to ensure effective alignment.
}
\paragraph{Deep CORAL}
\textcolor{black}{Unlike MMD, CORAL does not require a predefined kernel function, reducing the need for extensive hyperparameter tuning. Additionally, by aligning the second-order statistics (covariances) between domains, Deep CORAL can still effectively bridge domain gaps, particularly when they are not highly complex. Specifically, the core idea of Deep CORAL is to align the covariance matrices of features extracted from different domains. }Mathematically, the domain discrepancy is measured using the Frobenius norm of the difference between the source and target covariance matrices. The objective function of Deep CORAL is defined as:
\begin{equation}
    \label{eq:coral}
    \mathcal{L}_{\text{CORAL}} = \frac{1}{4d^2} \|C_s - C_t\|_F^2,
\end{equation}
where $C_s$ and $C_t$ are the covariance matrices of features from source and target domains respectively, and $d$ is the dimensionality.
\textcolor{black}{Compared to MMD, relatively few studies employed Deep CORAL, possibly due to the lack of rigorous theoretical proof. The approach proposed by Wang et al. \cite{wang2023sensor} utilized Deep CORAL to align sensor-level features across domains. This method was further enhanced through multi-graph alignment \cite{wang2023sea++}, where sequential graphs were constructed for both domains, and each graph was aligned independently. In this framework, graph-level alignment is weighted based on Deep CORAL computations, with larger weights assigned to graphs exhibiting greater domain discrepancies, emphasizing their alignment accordingly.} Besides aircraft engines, the metric-based methods were also widely adopted in bearing RUL prediction \cite{cheng2021transferable,mao2019predicting,xu2021deep,cao2021transfer,rathore2022rolling,wu2019weighted,zhuang2021temporal,zhu2020new,ding2021novel}.

These metric-based methods, including MMD and Deep CORAL, offer lower computational complexity and reduced risk of mode collapse, making them more stable and efficient compared to adversarial approaches. \textcolor{black}{However, they face key limitations that can impact their effectiveness in turbofan engine RUL prediction. One major challenge is metric selection, as the choice of similarity measure significantly affects adaptation performance. Selecting an inappropriate metric may lead to suboptimal feature alignment, causing the model to struggle with capturing degradation patterns across different engine conditions. Additionally, metric-based methods have difficulty handling highly nonlinear distribution shifts, especially when complex relationships between sensor features cannot be effectively captured by standard distance measures. This is particularly problematic in turbofan engine applications, where operational conditions vary dynamically, and degradation patterns exhibit irregular trends. As a result, these methods tend to perform better in scenarios with moderate distribution shifts, where feature variations remain relatively stable. To enhance their applicability to real-world turbofan engine RUL prediction, improvements such as adaptive metric learning or hybrid techniques combining metric-based and adversarial approaches may be necessary.}

\subsubsection{\textcolor{black}{Adversarial + Metric}}

\textcolor{black}{To leverage the advantages of both adversarial and metric-based methods, several studies \cite{fu2021deep,ye2022selective} have proposed hybrid approaches that combine these techniques to achieve better alignment.} For instance, A deep residual LSTM designed by Fu et al. \cite{fu2021deep} is composed of multiple LSTM layers and identity mapping. Then, they combined MK-MMD and DANN to reduce domain gaps. Similarly, a deep domain adaptative network proposed by Miao et al. \cite{miao2021deep} was based on a selective convolutional recurrent neural network to simultaneously optimize adversarial loss and metric loss with DANN and MK-MMD. \textcolor{black}{Zhao et al. \cite{zhao2024new} followed a similar hybrid framework, combining MK-MMD and DANN while incorporating a feature extractor with a residual separable convolutional module to enhance feature learning for cross-domain transfer.} Moreover, MMD and DANN have been used concurrently due to their practical applicability \cite{xie2024multidimensional47,ye2022selective}. The framework designed by Ye et al. \cite{ye2022selective} simultaneously integrates MMD and DANN to facilitate domain alignment while introducing a selective feature interaction module to optimize the selection of features for alignment.

These hybrid methods leverage the strengths of both adversarial and metric-based approaches, enabling them to handle significant domain variability while reducing the risk of mode collapse. \textcolor{black}{However, their increased computational complexity poses challenges for real-time turbofan engine RUL prediction, where timely failure estimation is crucial for predictive maintenance. High computational demands can delay inference, making these methods less practical for onboard diagnostics or real-time monitoring systems. Optimizing hybrid strategies to balance adaptation performance and efficiency is essential for their deployment in real-world turbofan engine applications.}

\subsubsection{Others}\color{black}
In addition to the commonly used DA methods, there are alternative approaches that can be applied in different settings. This section highlights two key strategies beyond the adversarial and metric-based methods: fine-tuning and mix-up.\paragraph{Fine-Tune}
\textcolor{black}{Fine-tuning aims to adapt a pre-trained model (typically trained on a source domain) to a target domain by retraining specific layers on the target data.} This allows a model initially trained on one engine type or operational condition to adjust to new environments with minimal labeled data. Zhang et al. \cite{zhang2018transfer} were the first to apply fine-tuning in turbofan engine RUL prediction by fine-tuning a bidirectional LSTM for the target domain, inspiring subsequent research \cite{chen2022multi,gribbestad2021transfer}. To enhance fine-tuning, DA techniques have also been incorporated \cite{chen2022rul,liu2024enhancing}. For instance, Chen et al. \cite{chen2022rul} proposed the DCNN-BiLSTM (TL) framework, which pretrains a model using MMD to reduce domain discrepancies before fine-tuning it on the target domain. However, integrating DA techniques in this manner requires access to both source and target domains during pretraining, which may not be practical for real-world applications. \textcolor{black}{To improve fine-tuning in a practical way, Seo et al. \cite{seo2024supervised} proposed combining LSTM and AdaBoost.} In this approach, the prediction error of each sample from the LSTM network is used to re-weight the samples, allowing the optimization to focus on challenging cases. Additionally, they developed hybrid models that integrate multiple models to further mitigate data distribution mismatches between source and target domains, making fine-tuning more effective.

Fine-tuning methods effectively capture general degradation patterns while adapting to domain-specific variations, making them particularly useful when the target domain has limited labels. By accelerating adaptation and reducing the need for training from scratch, they address data and time constraints in practical turbofan engine RUL prediction. \textcolor{black}{However, their reliance on target domain labels limits their applicability in real-world systems, where labeled degradation data is often scarce or unavailable, restricting their effectiveness in fully unsupervised adaptation scenarios.}

\paragraph{Mix-Up}
Mix-up was designed as a data augmentation strategy where new training samples are generated by linearly interpolating between pairs of samples and their labels. In DA, mix-up has been applied to enhance model robustness by encouraging better generalization across domains. Dong et al. \cite{dong2022dual} firstly introduced mix-up operations into a DANN-based framework to enrich the dataset. They proposed two levels of mix-up: time-series mix-up, which generates mixed samples within the source dataset for RUL prediction, and domain mix-up, which creates an intermediate domain by mixing samples from both the source and target domains. Since the target domain lacks labels, pseudo-labels were computed using an encoder pre-trained on the source domain. Inspired by this work, Furqon et al. \cite{furqon2024mixup} also employed mix-up for DA but without relying on traditional DA techniques like adversarial or metric-based methods. Their framework consists of three domains: source, intermediate, and target. In the source and target domains, two losses are computed—one for the original sample's prediction error and another for the mix-up sample's prediction error, with pseudo-labels used for target samples. Mix-up is performed by combining the features of two samples. In the intermediate domain, they apply mix-up at both the sample and feature levels by blending samples from the source and target domains. This approach achieves performance comparable to conventional DA techniques. 

\textcolor{black}{However, mix-up-based DA techniques lack a theoretical foundation to explain how they effectively reduce domain discrepancies. This gap in understanding is a significant limitation, particularly in complex tasks such as turbofan engine RUL prediction, where domain discrepancies can be large and subtle. For example, in turbofan RUL prediction, operational conditions, sensor configurations, and environmental factors can vary between domains, making it critical to understand how mix-up helps to bridge these differences. Thus, combining this technique with established DA techniques may offer a more feasible and theoretically grounded strategy for domain adaptation in turbofan engine RUL prediction.}

\color{black}\subsection{Alignment-based approaches}
\label{sec:alignment}
\textcolor{black}{This section summarizes the works based on where they handle distributional shifts, including marginal alignment and conditional alignment.
}
\subsubsection{Marginal Alignment}
\begin{figure}[htbp!]
    \centering
    \includegraphics[width = 1.\linewidth]{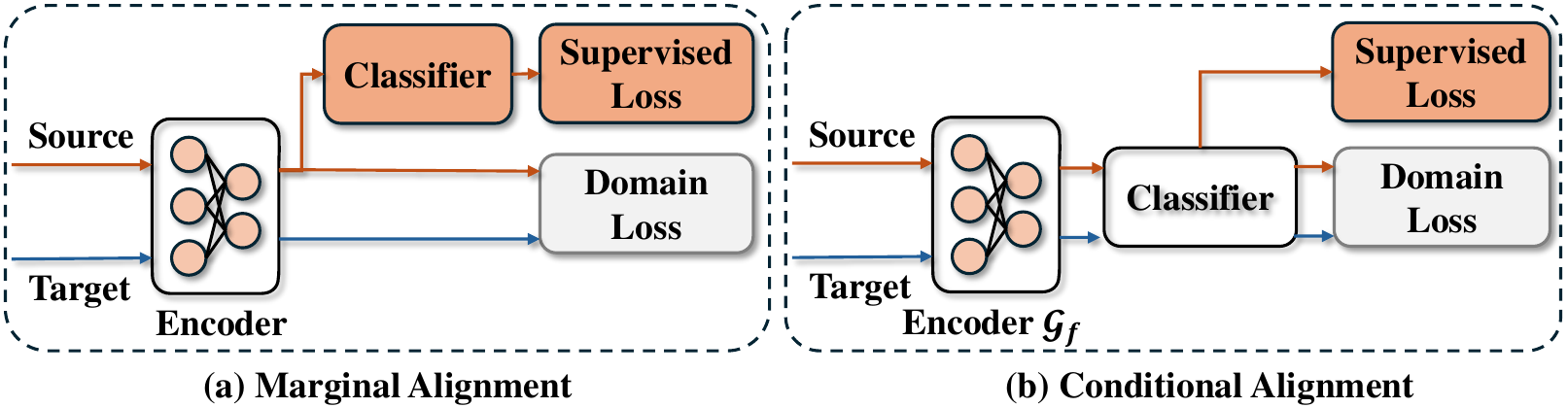}
    \caption{Marginal alignment vs. conditional alignment. \textcolor{black}{(a) Marginal alignment targets the overall distribution without considering labels. (b) Conditional alignment focuses on aligning distributions specific to different RUL stages.}}
    \label{fig:conditional}
\end{figure}

As illustrated in Fig. \ref{fig:conditional} (a), marginal distribution alignment focuses on aligning the overall feature distributions between the source and target domains. The goal is to ensure similarity in the marginal distributions of features across domains, without considering class labels or specific conditions. Many well-established DA techniques employ marginal alignment, including metric-based methods like MMD-based \cite{joseph2022lama,zhang2023variational,zhang2021transfer,wu2019weighted} and Deep CORAL-based frameworks \cite{wang2023sensor}, as well as adversarial-based methods such as DANN-based \cite{wu2022weighted,dong2022dual,li2022domain,nejjar2024domain} and ADDA-based frameworks \cite{ragab2020contrastive,ragab2020adversarial,wu2023privacy}.
Compared to conditional alignment, marginal alignment methods are relatively straightforward to implement and computationally efficient as they only focus on aligning the overall feature distributions without accounting for additional conditional dependencies. This makes them suitable for applications with limited computational resources. \textcolor{black}{However, marginal alignment does not explicitly capture the relationships between inputs and outputs, such as the connection between health indicators and RUL labels. As a result, it may lead to suboptimal alignment, especially in complex RUL prediction tasks, where different sensor readings correspond to varying degradation rates and operating conditions. This lack of explicit alignment between the input features and their corresponding outputs can hinder the model’s ability to accurately capture the nuanced patterns that drive RUL predictions, making it less effective in scenarios where the domain gap is large or the system's behavior is highly variable.}

\subsubsection{Conditional Alignment}

\textcolor{black}{To overcome the limitations of marginal alignment methods, conditional alignment has been developed, emphasizing the alignment of conditional distributions across domains.} A domain's distribution can be represented by its joint probability distribution $P(x,y)$, which decomposes into the marginal distribution $P(x)$ and conditional distribution $P(y|x)$. In the context of RUL prediction, aligning conditional distributions is crucial to ensure that samples corresponding to different stages of degradation are properly aligned \cite{li2020data,zou2022method,ding2021remaining}. \textcolor{black}{However, aligning conditional distributions presents unique challenges in regression tasks, where the absence of discrete classes makes it difficult to directly apply conventional alignment techniques. To address this, researchers have proposed innovative solutions, adapting alignment methods to effectively handle continuous labels and ensure meaningful alignment across domains.}

\textcolor{black}{The first approach involves transforming regression tasks into classification tasks to enable conditional DA for RUL prediction.} For instance, the entire lifecycle is segmented into multiple degradation stages, with each stage treated as a sub-category \cite{lyu2022remaining,he2023transferable36}. Conditional MMD is then applied to align these categories between source and target domains, using pseudo labels in the target domain where true labels are unavailable. A similar idea was adopted by Zhang et al. \cite{zhang2023variational}, who proposed a local weighted MMD method. This approach assigns weights based on the likelihood of samples belonging to specific categories, improving alignment compared to standard conditional MMD. Building on these ideas, Wen et al. \cite{wen2024unsupervised} employed an adversarial loss for alignment, introducing a sub-domain classification loss (with segmented RUL labels) to enhance conditional DA in RUL prediction. \textcolor{black}{Wen et al. \cite{wen2024unsupervised} further designed a sub-domain clustering network to assign sub-domain labels to each sample.} These labels are processed by a multi-linear conditioning module to generate new features, which are subsequently trained using DANN to account for conditional alignment effectively.

The above methods have demonstrated the effectiveness of achieving conditional alignment for turbofan engine RUL prediction. Although transforming regression tasks into classification by segmentation is straightforward, determining the optimal segmentation strategy and the appropriate number of sub-categories remains challenging. To address this issue, an alternative approach introduces weights during RUL prediction. Specifically, the weighted adversarial loss \cite{wu2022weighted} has been designed to prioritize samples with similar RULs across domains by adjusting weights based on their differences. Initially, an encoder is trained using source domain samples. Subsequently, the encoder is employed to generate pseudo labels for target domain samples. By computing the differences in RULs between samples from both domains, weights are assigned, with smaller weights assigned to samples exhibiting larger differences. This scheme ensures that samples with similar RUL values receive greater emphasis during training, enhancing the model's adaptation to the target domain.

In summary, compared to marginal alignment, conditional alignment matches feature distributions based on specific class labels, ensuring that task-relevant features are aligned between domains. This is particularly important in turbofan RUL prediction, where different sensors may exhibit distinct degradation patterns depending on the engine’s health state. However, focusing on conditional distributions inevitably increases computational complexity. For instance, dividing the degradation lifecycle into segments for alignment requires aligning each segment individually, \textcolor{black}{which not only adds to computational overhead but also increases the risk of convergence issues during training. This approach can become particularly challenging in large-scale turbofan engine datasets, where the volume of data and the need for efficient processing are paramount. The added complexity may result in longer training times, reduced scalability, and the potential for suboptimal performance, especially when dealing with a vast number of sensor readings and operational conditions.}

\subsection{Problem-based approaches}\color{black}
This section summarizes DA works specifically designed to tackle the unique challenges associated with turbofan engine RUL prediction. The challenges stem from various factors, including task characteristics, the domains in which these tasks are performed, the data utilized for modeling, and the sensors that gather information. Thus, the existing works are categorized into four levels: Task-level, Domain-level, Data-level, and Sensor-level, moving from coarse-grained to fine-grained perspectives and comprehensively covering both high-level and low-level challenges. Each category highlights a set of core problems that significantly influence the development of DA methods for turbofan engine RUL prediction.

\subsubsection{Task-level}
This level focuses on challenges that arise when predicting RUL alongside other tasks or transferring knowledge across tasks. Two core sub-categories are identified: multi-task optimization and task transfer.

\paragraph{Multi-Task Optimization}
\textcolor{black}{RUL prediction often requires models to simultaneously optimize multiple objectives, such as predicting RUL while assessing the State of Health (SOH).} By optimizing these tasks together, the model is encouraged to learn shared features, thereby reducing data shifts between domains. Zhang et al. \cite{zhang2023multi} proposed a multi-task learning-boosted method (MTLTrans) for cross-domain RUL prediction of turbofan engines. MTLTrans is built on a Transformer backbone with a hierarchical sharing structure, incorporating two auxiliary prognostic tasks: SOH and performance degradation prediction. The trade-off learning among these three tasks enhances the reliability of RUL predictions, making them more robust against data shifts. However, this work does not employ any alignment loss to explicitly address domain discrepancies, relying solely on optimizing multiple tasks to mitigate the impact of data shifts. \textcolor{black}{This approach may not fully eliminate domain discrepancies.} To address this limitation, Yan et al. \cite{yan2024feature} proposed a framework called FDMTCN, which integrates MMD into the multi-task optimization process, encompassing both SOH and RUL prediction. Additionally, they introduced a degradation correction loss to further regularize the prediction results.

\paragraph{Task-Transfer}
\textcolor{black}{In practice, models trained for one task often need to transfer knowledge to another task.} Task-transfer techniques enable models to leverage knowledge from related tasks, reducing the need for extensive retraining and enhancing adaptation to new scenarios. To facilitate this, Ruan et al. \cite{ruan2022fuzzy} developed a framework for transferring knowledge between different tasks, specifically fault classification and RUL prediction. The transfer process is straightforward; they first train a model on one task and then fine-tune it for the other task. However, RUL prediction is a regression problem, while fault classification is a classification problem, resulting in different label types—continuous for RUL prediction and discrete for fault classification. To overcome this issue, they employed a fuzzy membership function to convert discrete labels into continuous ones. \textcolor{black}{Notably, despite the promise of task-transfer techniques, few researchers have concentrated on this area. Therefore, increased attention to this direction could lead to significant advancements.}

\subsubsection{Domain-level}
This level focuses on the problems regarding domains, mainly including domain availability and domain-specific information that impacts adaptation performance in the target domain.

\paragraph{Domain Availability}
In most existing works, researchers typically assume the availability of a single source domain with labeled samples to mitigate domain discrepancies. However, this may not reflect real-world applications, where multiple source domains might be accessible, or no source domain is available due to data privacy concerns. In the first scenario, leveraging data from multiple source domains is essential for achieving better adaptation performance in the target domain. To address this, Du et al. \cite{du2024remaining} proposed a multi-source adversarial domain adaptation network. Specifically, they paired each source domain with the target domain and employed DANN to align each source-target domain pair. They designed a common feature extractor along with domain-specific feature extractors, which aim to learn unique features from each source domain. Additionally, they introduced a regularization loss to ensure that each domain feature extractor produces consistent RUL values for the target domain. This approach enables the model to learn more robust features for the target domain by enforcing uniform predictions across each source-target pair.

In cases where source data is unavailable due to privacy concerns, researchers must adapt a pretrained model solely to the target domain. To address this challenge, Wu et al. \cite{wu2023privacy} proposed a framework called SFDA-RUL, which focuses on training a target encoder using only the pretrained source model. The adaptation process involves an encoder and two predictors, all initialized with parameters from the source domain model. This process consists of two stages: first, the two predictors are trained to maximize their discrepancy by updating their parameters; in the second stage, the encoder is trained in an adversarial manner to minimize the discrepancy between the predictors.

\paragraph{Domain-Specific}
\begin{figure}[htbp!]
    \centering
    \includegraphics[width = 1.\linewidth]{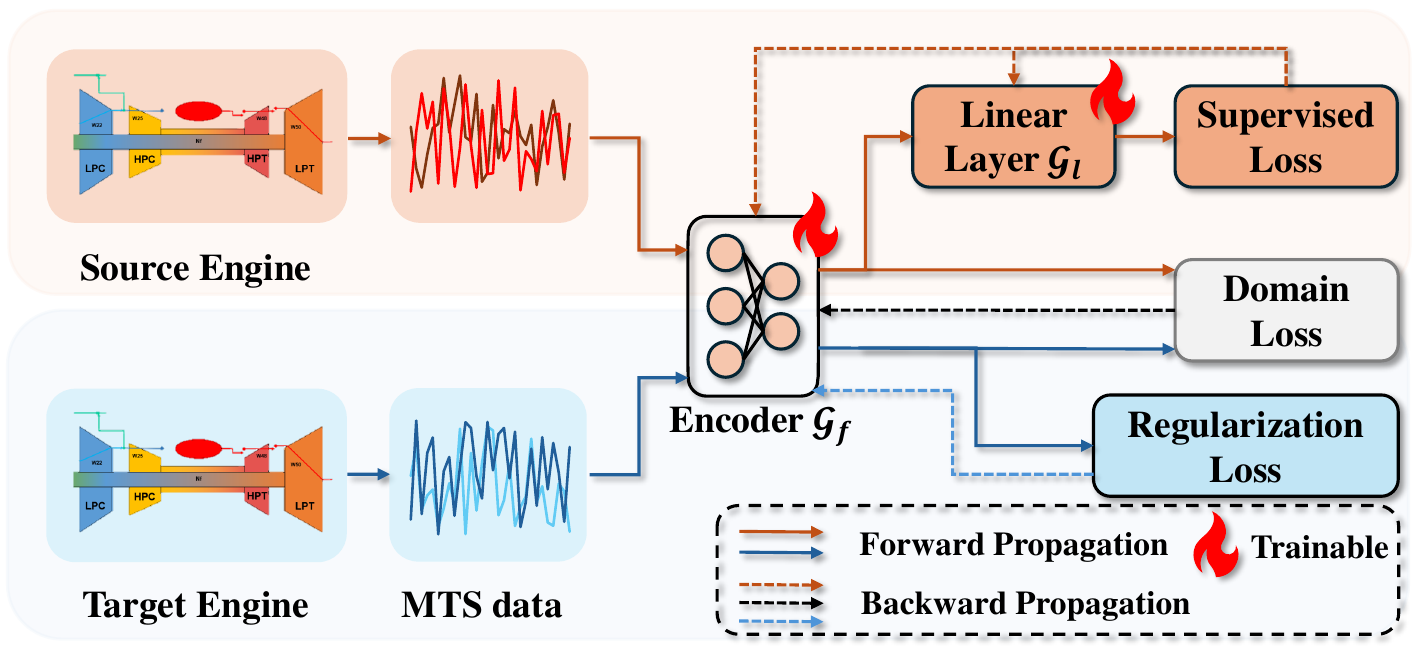}
    \caption{\textcolor{black}{DA methods for domain-specific information in RUL prediction, where the regularization loss is employed to preserve target domain information through reconstruction or contrastive techniques.}}
    \label{fig:recons}
\end{figure}
\textcolor{black}{Although existing DA methods have achieved effective alignment between two domains, they often focus solely on learning invariant features, neglecting domain-specific information within the target domain. This oversight can lead to the loss of crucial domain-specific details, ultimately impairing the performance of the adapted model on the target domain. To address this issue, several methods have been proposed to preserve target domain information as shown in Fig. \ref{fig:recons}}, where a regularization loss is introduced to retain key features. Two popular strategies for preserving target domain information are reconstruction loss and contrastive learning. 

The first strategy is reconstruction loss, which encourages the model to rebuild input data from its encoded representation, thereby retaining essential features. For instance, Joseph et al. \cite{joseph2022lama} proposed a reconstruction loss to help the model preserve discriminative information specific to the target domain. By minimizing reconstruction loss, the model learns to retain critical patterns in the data, ensuring that essential target-domain features are not lost during adaptation. The reconstruction method can effectively preserve fine-grained, domain-specific information, prompting the model to focus on detailed data patterns relevant to the target domain. However, the focus on detailed patterns can increase computational costs and may inadvertently reconstruct noisy information, degrading performance. \textcolor{black}{These methods also risk trivial solutions, where models memorize sensor data without capturing meaningful degradation patterns. This is particularly problematic in turbofan engines, where the degradation prediction requires precise signal interpretation.}

To address these limitations, contrastive learning has emerged as a complementary strategy. Unlike reconstruction, which focuses on rebuilding input samples at a fine-grained level, contrastive learning preserves discriminative information by maximizing the similarity between positive pairs (samples from the same class or domain) and minimizing the similarity between negative pairs (samples from different classes or domains). By applying this approach to the target domain, the model can learn embeddings that capture the inherent properties of target samples, thereby better retaining target domain-specific information. Due to its effectiveness, contrastive learning has led to growing interest in recent years, particularly for DA in RUL prediction. \textcolor{black}{For example, Contrastive Adversarial Domain Adaptation (CADA) proposed by Ragab et al. \cite{ragab2020contrastive} incorporated an InfoNCE loss to maximize mutual information between encoded representations of the target domain and the original inputs. Similarly, the contrastive loss was introduced by Dong et al. \cite{dong2022dual} to capture domain-specific features from the target domain.} Li et al. \cite{li2024pre} proposed a Pretraining-Enhanced Unsupervised Domain Adaptation (PEUDA) framework, where contrastive learning is used during pretraining by contrasting temporal and frequency augmentations. During adaptation, additional contrastive loss is applied to preserve target-domain information for improved DA performance. Although contrastive learning is effective, it comes with challenges. It requires carefully designed positive and negative pairs, making it more difficult to implement than reconstruction loss. Thus, developing techniques for the efficient selection of positive and negative pairs could further advance contrastive learning and enhance its ability to preserve target-domain information.

\textcolor{black}{The approaches discussed above primarily focus on retaining target-specific information to enhance performance in the target domain. Additionally, some methods also consider source-specific features. Unlike existing works that emphasize retaining target-specific information, methods focusing on source-specific features aim to eliminate the influence of source-specific information to facilitate the learning of better domain-shared features. For instance, Chen et al. \cite{chen2024transfer} identified a critical limitation in existing methods: they typically use all features to pre-train an RUL prediction model without addressing the negative impact of private features in the source domain on the target domain, which can hinder adaptation performance for turbofan engines. To address this, they proposed a framework called Transfer Regression Network-based Adaptive Calibration (TRNAC). TRNAC incorporates an adaptive calibration factor within its transfer regression network, enabling it to account for private features and automatically revise prediction errors based on the shared RUL predictor. While this approach presents an interesting direction, the challenge of mitigating the negative effects of source-specific information remains underexplored. Addressing this issue could offer a promising pathway to achieving more effective domain adaptation for turbofan engine RUL prediction.}

% \textcolor{black}{By inherently focus on reconstructing the input target data, these reconstruction-based methods  can maintain and leverage critical target-specific information, which is particularly valuable in turbofan engine RUL prediction, because it can help to capture subtle degradation trends and sensor variations in the target domain is crucial for accurate remaining useful life estimation. However, this is normally effective with sufficient target samples.  These methods may focus too much on preserving target-specific information, leading to potential overfitting. This can be problematic if the target domain is small or if the data quality is not consistent, as the model may capture noise instead of meaningful patterns.}
\subsubsection{Data-level}
This level addresses the intrinsic characteristics of the data used for turbofan engine RUL prediction, including uncertainty, time-varying properties, and incomplete degradation patterns within the data.
\paragraph{Uncertainty}
Engine degradation is inherently uncertain due to varying operational conditions, resulting in stochastic degradation patterns. DA models must be capable of capturing this uncertainty to provide reliable RUL predictions. To address uncertainty in cross-domain cases, Chen et al. \cite{chen2022multi} introduced a multi-granularity cross-domain temporal regression (MCDTR) network. This framework incorporates uncertainty quantification of predictions using a bootstrap strategy, enhancing the reliability and robustness of prognostic results for industrial turbofan engines. Following this, Cheng et al. \cite{cheng2024bayesian} proposed the BAPT framework, which employs a Bayesian neural network to quantify model uncertainty, providing interval estimates for RUL and improving model robustness. Additionally, they enhanced the standard transformer by introducing a multi-head ProbSparse attention module, focusing on dot-product operations highly correlated with degradation patterns. While these approaches emphasize the importance of uncertainty in turbofan engine RUL prediction, integrating uncertainty into the alignment process remains under-explored. Further research is required to develop more effective methods that seamlessly combine alignment strategies with uncertainty such as uncertainty alignment, leading to improved DA performance for turbofan engines.

\paragraph{Time-Varying}
Turbofan engines operate under dynamic conditions, with degradation patterns evolving over time. Accurate RUL predictions require DA techniques that can address these time-varying characteristics by aligning temporal dependencies across source and target domains. To tackle the challenge of aligning time-varying distributions in time-series data, Xiang et al. \cite{xiang2024micro} proposed a micro-transfer learning framework based on MCLSTM, called MC-DANN. The framework leverages a multi-cellular LSTM, which consists of multiple sub-cell units with distinct update modes, enabling the model to capture and differentiate between varying distributions within the input data. To align information across domains effectively, the sub-cells (micro-level) are aligned using DANN, while global features (macro-level) are also aligned through DANN. This dual-level alignment ensures both local and global temporal dependencies are matched between domains. Although the current method demonstrates the potential of accounting for time-varying properties, further exploration is needed, as the time-varying property is an important characteristic in turbofan engines. One key area involves improving scalability when dealing with complex temporal patterns, as more intricate update modes may increase computational overhead. Additionally, integrating advanced attention mechanisms could enhance the alignment of temporal dependencies, especially when handling long sequences. These extensions would further refine DA methods, making them better suited for real-world applications in turbofan engine RUL prediction.

\paragraph{Incomplete}

\begin{figure}[htbp!]
    \centering
    \includegraphics[width = .8\linewidth]{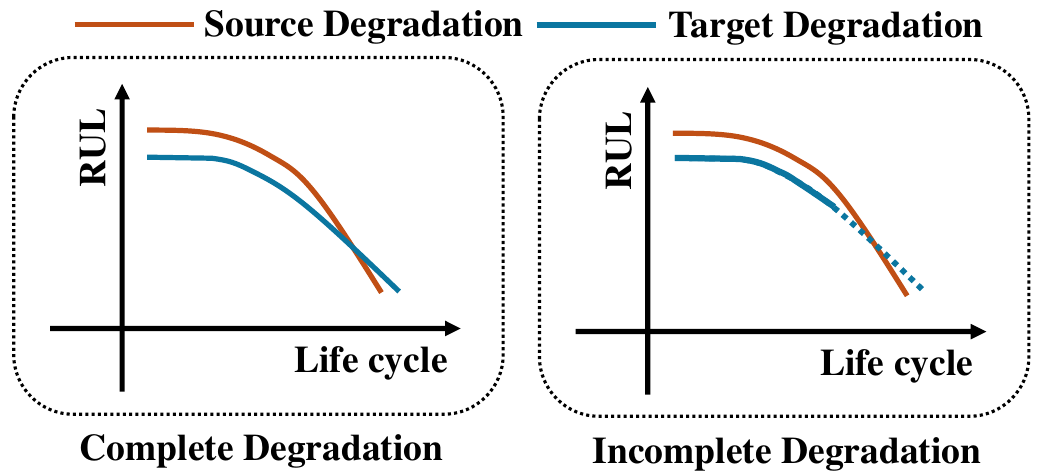}
    \caption{\textcolor{black}{Complete vs. incomplete degradation in turbofan engine RUL prediction. In real-world scenarios, target domains often contain samples primarily from healthy stages, with few or no samples from degradation stages, posing challenges for accurate prediction.}
}
    \label{fig:degra}
\end{figure}

In real-world RUL prediction scenarios, acquiring degradation data is often challenging, leading to incomplete life cycles in the target domain \cite{zhang2021transfer, siahpour2022novel, cheng2022two}. Specifically, the target domain typically contains only health-stage data, with little to no degradation-stage data, as shown in Fig. \ref{fig:degra}. This absence of near-failure features poses a significant obstacle to DA \cite{krokotsch2020novel}. To address this issue for practical applications, several strategies have been proposed. One solution involves categorizing the life cycle into health and degradation states, which can then be aligned respectively to address the incomplete problem. The framework designed by Zhang et al. \cite{zhang2021transfer} pioneered this approach by aligning data within each state. For the health state, domain fusion is assumed, ensuring that both domains share the same feature space. For the degradation state, alignment is designed based on the assumption that all degradation trends follow a consistent direction from the health state. A key challenge in such solution is identifying the degradation stage for alignment across domains. To address this, the concept of the first degradation point was introduced \cite{li2021degradation}, aiming to align the degradation stages across domains and ensure cycle consistency. Beyond state-based segmentation, some methods divide the entire life cycle into multiple segments for more precise alignment \cite{he2023transferable36}. These methods segment RUL values and compute two alignment losses—class-wise and marginal-wise. Additionally, a manifold regularization loss is applied to handle incomplete target data, guiding the model to predict RUL values closer to failure.

\textcolor{black}{While such segmentation methods show promising results, they depend on the availability of clear health and degradation stages in the target domain, which may not always be present.} To overcome this limitation, some frameworks aim to detect incomplete parts automatically. For instance, Siahpour et al. \cite{siahpour2022novel} proposed a two-stage training process. In the first stage, an encoder is trained to model relationships between inputs and RUL labels, capturing both minor and severe degradation trends. In the second stage, given the lack of sufficient degradation data in the target domain, the framework employs consistency-based regularization using pretrained features from the first stage to align source and target features effectively.

\subsubsection{Sensor-level}

Due to the large volume of turbofan engines, multiple sensors are typically deployed to enhance status detection. Here, exploring sensor-related problems has become a crucial research direction. Existing works can be broadly categorized into two approaches: those focusing on leveraging sensor-level properties for better alignment and those addressing sensor-related issues, such as sensor malfunction.

\paragraph{Sensor Characteristic}
Given the multi-sensor nature, with each sensor playing distinct roles in status detection, addressing sensor-level characteristics is crucial for mitigating domain discrepancies \cite{wang2023sensor, chen2023aero, wang2023sea++}. Xie et al. \cite{xie2024multidimensional47} proposed a sensor channel attention mechanism to effectively fuse multi-source information by computing the importance of each channel for the final representations. However, their approach focuses on sensor-level representation learning while neglecting explicit sensor-level alignment. \textcolor{black}{To address this gap, Wang et al. \cite{wang2023sensor} designed both sensor feature alignment and sensor correlation alignment, using Deep CORAL to reduce domain shifts. Additionally, they proposed a Graph Neural Network (GNN)-based encoder \cite{zhu2023rgcnu, wang2024fully} to model inter-sensor correlations.} Building on this, these authors further introduced multi-graph alignment and higher-order statistics alignment to enhance sensor-level alignment for improved DA performance in turobofan engines \cite{wang2023sea++}. \textcolor{black}{Meanwhile, the problem of potential inconsistencies in sensors was addressed by Chen et al. \cite{chen2023aero} by considering trends such as increasing or decreasing values. To align these trends and prevent inconsistencies, they introduced a self-adaptive dynamic clustering method, generating two clusters for sensors and aligning them between domains.}

\paragraph{Sensor Malfunction}
Due to mechanical damage or software errors, sensor malfunction may occur, rendering partial sensor channels unavailable and affecting the performance of existing DA techniques. To address this issue, Li et al. \cite{li2022remaining39} proposed a framework that employs a shared feature extractor to learn invariant sensor-level features. Specifically, the shared feature extractor is used to extract features from each sensor across different engines. DANN is then applied both across sensors and across engines to align the extracted features. Two discriminators are introduced: one to determine the source of the signal (which sensor it comes from) and another to identify the engine generating the signal. By employing these discriminators, the framework encourages the feature encoder to learn sensor-engine-invariant features. This approach helps the model remain robust to missing or malfunctioning sensor channels, mitigating their negative impact on performance.

\textcolor{black}{Compared to other RUL-related challenges, sensor-related issues still remain under-explored, particularly in leveraging sensor characteristics for improved alignment and addressing sensor malfunction problems.} Given the crucial role of sensors in monitoring engine health, future research should focus more on these areas to enhance DA in turbofan engines. \textcolor{black}{Advancing methods for sensor-level alignment and developing robust solutions to handle sensor malfunctions can significantly improve the reliability and performance of RUL prediction models in real-world applications.}

\color{black}

\begin{table}[]
\centering
\caption{Details of C-MAPSS dataset.}
\label{tab:cmapss}
\begin{tabular}{l|cccc}
\toprule
\toprule
Sub-dataset                    & FD001 & FD002 & FD003 & FD004 \\ \midrule
\# Engine units for training        & 100   & 260   & 100   & 249   \\ \midrule
\# Engine units for testing       & 100   & 259   & 100   & 248   \\ \midrule
\# Training samples        & 17731 & 48558 & 21220 & 56815 \\ \midrule
\# Testing samples         & 100   & 259   & 100   & 248   \\ \midrule
%\# Max life spans (cycles) & 362   & 378   & 512   & 128   \\ \hline
\# Operating conditions    & 1     & 6     & 1     & 6     \\ \midrule
\# Fault modes             & 1     & 1     & 2     & 2     \\ \bottomrule\bottomrule
\end{tabular}
\end{table}

\begin{table}[]
\centering
\caption{Details of N-CMAPSS dataset.}
\label{tab:ncmapss}
\begin{tabular}{l|ccc}
\toprule
\toprule
Sub-dataset                    & DS01 & DS02 & DS03 \\ \midrule
\# Engine units for training        & 6   & 6   & 9    \\ \midrule
\# Engine units for testing       & 4   & 3   & 6     \\ \midrule
\# Training samples        & 48773 & 52341 & 55272  \\ \midrule
\# Testing samples         & 27157   & 12391   & 42222      \\ \midrule
\# Affected components    & HPT     & HPT, LPT     & HPT, LPT          \\ \midrule
\# Fault modes             & 1     & 2     & 1         \\ \bottomrule\bottomrule
\end{tabular}
\end{table}

\section{Comparative Analysis}
\subsection{Datasets}
\textcolor{black}{We employ the C-MAPSS and N-CMAPSS benchmark datasets to evaluate the performance of state-of-the-art methods.} The C-MAPSS dataset \cite{4711414} comprises operational data from four distinct turbofan engines. As detailed in Table~\ref{tab:cmapss}, each engine operates under unique conditions and exhibits specific fault modes. This dataset features data from 21 sensors strategically placed to monitor engine health. We followed the data preparation from \cite{xu2023hybrid} for preprocessing. The refined dataset includes data from 14 selected sensors due to the constant values of the remaining sensors \cite{wang2023local,jin2022position}, with the labels indicating the remaining useful life of the engines. The N-CMAPSS dataset~\cite{data6010005} captures the run-to-failure trajectories of turbofan engines. Unlike the C-MAPSS dataset, which is limited to standard cruise phase conditions, N-CMAPSS encompasses simulations of complete flights, including climb, cruise, and descent phases. Additionally, it enhances the fidelity of degradation modeling. \textcolor{black}{These enhancements enable N-CMAPSS to more accurately reflect the complex factors present in real systems.} We utilize datasets DS01, DS02, and DS03, which record data from 20 channels, for our experiments as outlined in Table~\ref{tab:ncmapss}. The preprocessing follows the methodology described in \cite{mo2022multi}.

\subsection{Experimental Setting}

We employ the Root Mean Square Error (RMSE) and Score metrics to evaluate our model, as outlined in \cite{ragab2020contrastive}. Lower values of these indicators signify better model performance.

The RMSE metric is defined as follows:
\begin{equation}
RMSE = \sqrt{\frac{1}{N}\sum_{i=1}^N (y_i-\widehat{y_i})^2},
\label{eq:rmse}
\end{equation}
where $\widehat{y_i}$ represents the estimated RUL, and $y_i$ denotes the actual RUL. RMSE treats early and late predictions of RUL equally. However, in prognostics, late predictions of RUL can be more harmful. To address this, we use the Score metric, which applies a heavier penalty for late predictions, thus reflecting the urgency and importance of accurate late-stage prognostics. The current Score metric increases monotonically with the quantity of test samples, indicating a lack of normalization and making it challenging to compare across different datasets fairly. To address this issue, we are using the average Score. The formulation of the average Score metric is as follows:

\begin{equation}
Score_i =\left\{\begin{aligned}
 e^{-\frac{\widehat{y_i}-y_i}{13}}-1; \widehat{y_i} < y_i, \\
 e^{\frac{\widehat{y_i}-y_i}{10}}-1; \widehat{y_i} > y_i, 
\end{aligned}\right.
\label{eq:score_i}
\end{equation}

\begin{equation}
Score =\sum_{i=1}^N Score_i.
\label{eq:avg_score}
\end{equation}

\begin{equation}
Avg Score =\frac{Score}{N}.
\label{eq:avg_score}
\end{equation}

For better comparison, we ranked all approaches based on their RMSE and average Score under various scenarios and listed the average ordinal positions for each approaches.

% Comparing the effectiveness of Transfer Learning approaches against traditional methods

% Actions:
% \begin{itemize}
%     \item Convert ADATIME to fit for RUL prediction task: Change backbone, dataloader,  loss function, and evaluation metrics
%     \item Run experiments of multiple RUL Baselines on CMAPSS dataset
%     \item Provide results and key findings (Best method, worst method, easiest method, common failures) 
%     \item Summarize the results in Tables, or nice graphs, such as radar chart 
% \end{itemize}

\subsection{Benchmarking Results}

\begin{table*}[htbp]\color{black}
  \centering
  \scriptsize
  \caption{Benchmarking results on CMAPSS for RMSE (Bold indicates the best result, and underline indicates the second-best result).}
    \begin{adjustbox}{width = 1\textwidth,center}

    \begin{tabular}{l|c|c|c|c|c|c|c|c|c|c|c|c|c|c}
    \toprule
    \toprule
    Methods & F1$\to$F2 & F1$\to$F3 & F1$\to$F4 & F2$\to$F1 & F2$\to$F3 & F2$\to$F4 & F3$\to$F1 & F3$\to$F2 & F3$\to$F4 & F4$\to$F1 & F4$\to$F2 & F4$\to$F3 & Avg.  & Rank \\
    \midrule
    Source & \underline{17.20} & 51.49 & 41.04 & 16.43 & 36.41 & 37.96 & 60.82 & 50.13 & 24.37 & 41.19 & 40.53 & 19.40 & 36.41 & 6.42 \\
    DDC   & \textbf{16.02} & 34.92 & 27.35 & \textbf{16.27} & 23.19 & 22.71 & 25.16 & \underline{22.74} & \underline{19.94} & \textbf{21.91} & 25.50 & \underline{19.00} & 22.89 & \underline{3.17} \\
    CORAL & 23.31 & 32.69 & 42.56 & 17.39 & 26.50 & 34.80 & 33.33 & 32.29 & 27.77 & 36.41 & \underline{23.23} & 19.29 & 29.13 & 6.00 \\
    HoMM  & 19.70 & 60.04 & 42.62 & 19.24 & 45.00 & 42.77 & 64.90 & 61.00 & 28.74 & 48.38 & 37.08 & 21.96 & 40.95 & 8.33 \\
    DANN  & 21.96 & 27.79 & 30.18 & 19.44 & 26.64 & 27.25 & 43.34 & 22.84 & 24.84 & 31.71 & 23.40 & 21.19 & 26.71 & 5.42 \\
    AdvSKM & 17.30 & 31.12 & 35.29 & 20.50 & 39.89 & 30.65 & 27.42 & 29.24 & 25.26 & 26.43 & 32.02 & 21.88 & 28.08 & 5.75 \\
    ConsDANN & 17.92 & \underline{20.85} & 22.83 & 16.37 & \underline{20.28} & 26.19 & \underline{22.92} & 23.90 & 21.86 & 24.66 & \textbf{21.19} & 25.80 & \underline{22.06} & 3.42 \\
    ADARUL & 17.44 & 21.01 & \underline{18.66} & 21.39 & 21.99 & \textbf{20.52} & 25.35 & 36.19 & \textbf{19.14} & \underline{24.55} & 26.32 & \textbf{18.54} & 22.59 & 3.42 \\
    CADA  & 22.95 & \textbf{19.41} & \textbf{18.61} & \underline{16.30} & \textbf{18.00} & \underline{20.53} & \textbf{22.45} & \textbf{18.95} & 21.65 & 29.81 & 34.41 & 19.77 & \textbf{21.90} & \textbf{3.08} \\
    \bottomrule
    \bottomrule
    \end{tabular}%
    \end{adjustbox}
  \label{tab:bench_CMPS_RMSE}%
\end{table*}%

\begin{table*}[htbp]\color{black}
  \centering
  \scriptsize
  \caption{Benchmarking results on CMAPSS for Score (Bold indicates the best result, and underline indicates the second-best result).}
    \begin{adjustbox}{width = 1\textwidth,center}
    \begin{tabular}{l|c|c|c|c|c|c|c|c|c|c|c|c|c|c}
    \toprule
    \toprule
    Methods & F1$\to$F2 & F1$\to$F3 & F1$\to$F4 & F2$\to$F1 & F2$\to$F3 & F2$\to$F4 & F3$\to$F1 & F3$\to$F2 & F3$\to$F4 & F4$\to$F1 & F4$\to$F2 & F4$\to$F3 & Avg.  & Rank \\
    \midrule
    Source & 22.63 & 176.25 & 95.15 & 4.92  & 57.04 & 68.90 & 35936.22 & 15712.96 & 70.06 & 1215.38 & 313.58 & 15.90 & 4474.08 & 6.67 \\
    DDC   & 40.60 & 278.46 & 39.92 & \underline{5.73} & 21.38 & \textbf{11.97} & 44.55 & 37.28 & \underline{14.15} & \textbf{21.46} & 60.32 & 66.09 & 53.49 & 4.58 \\
    CORAL & 23.95 & 45.16 & 105.39 & 8.56  & 32.27 & 48.07 & 221.49 & 160.38 & 64.70 & 857.41 & 139.82 & 17.63 & 143.74 & 6.00 \\
    HoMM  & 31.26 & 426.46 & 107.60 & 17.43 & 133.35 & 133.62 & 15395.27 & 4069.94 & 259.58 & 1761.07 & 625.67 & 21.70 & 1915.25 & 8.50 \\
    DANN  & 18.41 & 38.03 & 32.04 & 12.11 & 38.88 & 21.25 & 468.30 & 732.44 & 90.28 & 230.57 & 229.51 & 16.39 & 160.69 & 5.67 \\
    AdvSKM & 13.83 & 49.03 & 147.77 & 15.11 & 112.99 & 42.22 & 50.11 & \underline{22.38} & 25.88 & \underline{56.26} & \underline{26.19} & 19.09 & 48.41 & 5.08 \\
    ConsDANN & \underline{7.31} & \underline{18.64} & 14.01 & 6.02  & \underline{19.75} & 22.84 & \underline{23.38} & 27.11 & 37.18 & 70.41 & \textbf{13.34} & 44.81 & \underline{25.40} & 3.42 \\
    ADARUL & \textbf{7.13} & 22.15 & \underline{11.53} & 13.38 & 21.14 & \underline{13.60} & 51.09 & 35.32 & \textbf{8.34} & 73.45 & 40.31 & \textbf{11.33} & 25.73 & \underline{3.17} \\
    CADA  & 7.70  & \textbf{17.10} & \textbf{10.10} & \textbf{4.92} & \textbf{11.14} & 13.79 & \textbf{21.15} & \textbf{17.21} & 17.19 & 56.41 & 32.12 & \underline{11.88} & \textbf{18.39} & \textbf{1.92} \\
    \bottomrule
    \bottomrule
    \end{tabular}%
    \end{adjustbox}
  \label{tab:bench_CMPS_score}%
\end{table*}%
\begin{table}[htbp]\color{black}
  \centering
  \caption{Benchmarking results on N-CMAPSS for RMSE.}
  \begin{adjustbox}{width = .5\textwidth,center}
    \begin{tabular}{lcccccccc}
    \toprule
    \toprule
    Methods & D1$\to$D2 & D1$\to$D3 & D2$\to$D1 & D2$\to$D3 & D3$\to$D1 & D3$\to$D2 & Avg.  & Rank \\
    \midrule
    Source & 25.86 & \textbf{9.40} & 27.20 & 25.21 & 36.45 & 19.78 & 23.98 & 7.00 \\
    DDC   & 12.43 & 11.80 & 24.56 & 16.73 & 25.19 & \underline{14.60} & 17.55 & 4.33 \\
    CORAL & 13.00 & 10.82 & 24.65 & 16.40 & 35.95 & 17.42 & 19.71 & 5.50 \\
    HoMM  & 12.53 & 11.64 & 25.41 & 17.05 & 36.92 & 16.65 & 20.03 & 6.67 \\
    DANN  & 12.64 & 11.21 & 25.21 & \underline{14.08} & \underline{17.35} & 17.03 & \underline{16.25} & \underline{4.17} \\
    AdvSKM & \underline{12.31} & 11.51 & 25.21 & 16.79 & 25.09 & 16.38 & 17.88 & 4.67 \\
    ConsDANN & \textbf{11.98} & 10.63 & \underline{21.46} & \textbf{13.48} & \textbf{16.24} & 20.70 & \textbf{15.75} & \textbf{2.83} \\
    ADARUL & 27.51 & \underline{9.94} & \textbf{19.34} & 14.92 & 32.67 & 17.19 & 20.26 & 4.33 \\
    CADA  & 29.10 & 12.20 & 24.79 & 15.33 & 30.48 & \textbf{7.22} & 19.85 & 5.50 \\
    \bottomrule
    \bottomrule
    \end{tabular}%
    \end{adjustbox}
  \label{tab:bench_NCMPS_RMSE}%
\end{table}%

\begin{table}[htbp]\color{black}
  \centering
  \caption{Benchmarking results on N-CMAPSS for Score.}
  \begin{adjustbox}{width = .5\textwidth,center}
    \begin{tabular}{lcccccccc}
    \toprule
    \toprule
    Methods & D1$\to$D2 & D1$\to$D3 & D2$\to$D1 & D2$\to$D3 & D3$\to$D1 & D3$\to$D2 & Avg.  & Rank \\
    \midrule
    Source & 67.84 & \textbf{1.19} & 12.08 & 50.91 & 29.47 & 4.48  & 27.66 & 7.33 \\
    DDC   & \textbf{2.19} & 2.66  & 9.55  & 4.93  & 9.52  & \underline{2.75} & 5.27  & 4.17 \\
    CORAL & 2.65  & 2.00  & 9.81  & 4.73  & 28.11 & 3.43  & 8.46  & 5.67 \\
    HoMM  & 2.44  & 2.49  & 10.60 & 5.49  & 30.86 & 3.13  & 9.17  & 6.67 \\
    DANN  & \underline{2.31} & 2.20  & 10.15 & 3.33  & \underline{4.43} & 3.55  & \underline{4.33} & 4.33 \\
    AdvSKM & 2.50  & 2.34  & 10.26 & 5.04  & 9.65  & 3.00  & 5.46  & 5.33 \\
    ConsDANN & 2.57  & 1.96  & \underline{6.62} & \textbf{2.39} & \textbf{3.95} & 5.07  & \textbf{3.76} & \underline{3.67} \\
    ADARUL & 28.44 & \underline{1.29} & \textbf{4.86} & 4.37  & 19.90 & 3.40  & 10.38 & 4.33 \\
    CADA  & 28.08 & 1.81  & 8.40  & \underline{2.87} & 16.04 & \textbf{0.84} & 9.67  & \textbf{3.50} \\
    \bottomrule
    \bottomrule
    \end{tabular}%
    \end{adjustbox}
  \label{tab:bench_NCMPS_score}%
\end{table}%
\textcolor{black}{We selected SOTA DA methods for RUL predictions, including DDC~\cite{tzeng2014deep}, Deep CORAL~\cite{sun2017correlation}, HoMM \cite{chen2020homm}, DANN \cite{Ganin2017Domain}, AdvSKM \cite{liu2021adversarial}, ConsDANN \cite{siahpour2022novel}, ADARUL~\cite{ragab2020adversarial}, and CADA~\cite{ragab2020contrastive}. Additionally, we compared these methods with a baseline approach, referred to as "Source Only" (Source), where the model trained on the source domain is directly applied to the target domain. To ensure a fair comparison, all methods were implemented with the same feature extractor and predictor architecture, and each method was run five times to report average performance. The feature extractor consists of a three-layer 1D CNN, where each convolutional layer is followed by a LeakyReLU activation function. After the three convolutional layers, a max-pooling layer with a kernel size of 5 is applied, and the output is flattened to form a feature vector, which serves as input to the RUL prediction model. For optimal performance across all methods, we followed the protocol outlined in previous work~\cite{ragab2023adatime} and used source risk to select the best hyperparameters. Specifically, we select the model that achieves the lowest mean squared error on a validation set from the source domain. This approach ensures that hyperparameter tuning is performed independently of target domain labels, maintaining the integrity of DA by preventing information leakage from the target domain.
}

\textcolor{black}{The results, presented in Tables \ref{tab:bench_CMPS_RMSE} and \ref{tab:bench_CMPS_score} for the C-MAPSS dataset and Tables \ref{tab:bench_NCMPS_RMSE} and \ref{tab:bench_NCMPS_score} for the N-CMAPSS dataset, include RMSE and Score metrics. Overall, CADA, ADARUL, and ConsDANN demonstrated superior performance, likely due to their specialized design for turbofan engine RUL prediction. Additionally, simpler methods like DDC and DANN achieved relatively good performance, as their straightforward frameworks make them easily adaptable to this task. }

Interestingly, in specific cross-domain cases, such as F1$\to$F2 for RMSE in C-MAPSS and D1$\to$D3 for RMSE in N-CMAPSS, the source-only approach achieved better performance than other SOTA methods. This can be attributed to two key factors. First, in these cases, the domain shift is minimal, meaning the distributions of the source and target domains are highly similar. In such scenarios, DA techniques may provide little to no benefit, as the alignment process becomes redundant. Second, noise in the target domain can adversely impact the adaptation process. While DA techniques aim to align the target distribution with the source, they may inadvertently align noise in the target data, leading to negative transfer and ultimately poorer performance compared to the source-only approach. \textcolor{black}{These observations highlight the importance of carefully evaluating the need for DA techniques based on the extent of domain shifts and data quality to avoid unnecessary complexities or performance degradation.}
% \renewcommand{\arraystretch}{1.3}

% \vspace{-0.4cm}
% \begin{figure}[htbp!]
%     \centering
%     \includegraphics[width = \linewidth]{Figures/cmapss_radar_rmse.pdf}
%     \caption{Comparative RMSE Analysis in a Radar Chart for C-MAPSS Dataset. Each axis represents a scenario. Small enclosed area indicates superior performance.}
%     \label{fig:cmapss_radar}
% \end{figure}

% \vspace{-0.5cm}

% \begin{figure}[htbp!]
%     \centering
%     \includegraphics[width = \linewidth]{Figures/ncmapss_radar_rmse.pdf}
%     \caption{Comparative Avg Score Analysis in a Radar Chart for N-CMAPSS Dataset. Each axis represents a scenario. Small enclosed area indicates superior performance.}
%     \label{fig:ncmapss_radar}
% \end{figure}

\color{black}\subsection{Complexity Analysis}

\begin{table}[htbp]\color{black}
  \centering
  \caption{Running time of SOTA methods.}
  \begin{adjustbox}{width = .24\textwidth,center}
    \begin{tabular}{l|cc}
    \toprule
    \toprule
    \multirow{2}[2]{*}{Methods} & \multicolumn{2}{c}{Time/s} \\
          & CMAPSS & N-CMAPSS \\
    \midrule
    DDC   & 339   & 588 \\
    CORAL & 300   & 546 \\
    HoMM  & 429   & 4992 \\
    DANN  & \underline{276}   & \textbf{504} \\
    AdvSKM & 690   & 1104 \\
    ConsDANN & 321   & 594 \\
    ADARUL & \textbf{270}   & \underline{516} \\
    CADA  & 480   & 984 \\
    \bottomrule
    \bottomrule
    \end{tabular}%
    \end{adjustbox}
  \label{tab:running_time}%
\end{table}%`

\begin{table}[htbp]\color{black}
  \centering
  \caption{Model complexity of non-adversarial and adversarial methods.}
  \begin{adjustbox}{width = .35\textwidth,center}
    \begin{tabular}{l|cc}
    \toprule
    \toprule
    Methods & FLOPs & No. Parameters \\
    \midrule
    Non-Adversarial & 25,837,568 & 16,400 \\
    Adversarial & 25,837,568+4,227,072 & 16,400+8,400 \\
    \bottomrule
    \bottomrule
    \end{tabular}%
    \end{adjustbox}
  \label{tab:flops}%
\end{table}%

Model complexity analysis is crucial for assessing the practicality of DA techniques for turbofan engine RUL prediction. Typical metrics include the number of floating-point operations (FLOPs) and model parameters. However, since all methods share the same backbone architecture, direct comparisons of these indicators can be challenging. This is particularly relevant for adversarial-based methods, such as DANN, ConsDANN, ADARUL, and CADA, which require additional discriminators to perform the adversarial process. Therefore, we compare FLOPs and number of model parameters specifically between non-adversarial and adversarial methods. Additionally, for a more comprehensive evaluation, we also assess the running time of all methods. To compare running time, we implemented each method on an RTX A4000 GPU for 100 epochs. Since C-MAPSS and N-CMAPSS datasets involve different numbers of cross-domain cases, we conducted experiments across all cases and reported the average running time.

Table \ref{tab:flops} presents the comparisons for FLOPs and model parameters. The results show that the inclusion of additional discriminators in adversarial methods increases model complexity during the training phase. However, this additional complexity is limited to training, as the discriminators are not used during inference. Regarding running time in Table \ref{tab:running_time}, we observe that ADARUL and DANN require less time for training, indicating their efficiency. Additionally, training on the N-CMAPSS dataset takes longer than on C-MAPSS due to the larger number of samples in N-CMAPSS. Among the evaluated methods, HoMM demonstrates poor scalability, with its training time increasing nearly tenfold when transitioning from C-MAPSS to N-CMAPSS. These findings highlight the trade-offs between complexity and efficiency for different DA techniques.

\color{black}
\color{black}\section{Practical Applications of DA Techniques for RUL Prediction in Turbofan Engines}
The integration of DA techniques for RUL prediction in the context of turbofan engines has significant implications for the aviation industry. The following subsections discuss industry-specific considerations, data-related challenges, practical implementation issues, and potential real-world applications to illustrate the value and limitations of these methods.
\subsection{Industry Standards and Maintenance Practices}
\textcolor{black}{In the aviation sector, Maintenance, Repair, and Overhaul (MRO) processes adhere to strict regulatory standards set by organizations such as the Federal Aviation Administration (FAA) and the European Union Aviation Safety Agency (EASA).} These standards mandate regular inspections, strict monitoring of engine performance, and precise failure prognostics to ensure flight safety and operational efficiency. Although data-driven methods have shown promising results in this area, they face difficulties when transferred across different operational settings or between different engine models.

\textcolor{black}{DA offers a promising solution by enabling models trained on existing data (e.g., historical data from one fleet or engine type) to generalize effectively to new environments (e.g., a different fleet, varying flight conditions, or new engine configurations).} For instance, DA can help align degradation patterns between older engine models and newer variants, ensuring consistent RUL predictions without requiring extensive labeled data from each new configuration. \textcolor{black}{This capability aligns with industry efforts to reduce the reliance on expensive and time-consuming data collection, paving the way for predictive maintenance that complies with aviation regulations while minimizing downtime.}

\subsection{Data Availability}
Data availability is a critical concern for RUL prediction in turbofan engines. In practice, only a limited amount of run-to-failure data is available, as most engines are removed from service before catastrophic failure occurs. Additionally, there are variations in data characteristics across different airlines, engines, and operational conditions, resulting in substantial domain shifts. DA methods can address these challenges by enabling the use of abundant data from similar but non-identical domains, such as simulated datasets or historical records from different fleets, to train models that can perform well on the target domain.

\subsection{Implementation Challenges and Practical Considerations}
In this section, we discuss the implementation challenges of applying existing DA techniques in real-world applications. Notably, corresponding solutions are outlined in the subsequent section for future research directions.

\subsubsection{Data Privacy and Security}
A major barrier to the deployment of DA in turbofan engine RUL prediction is the sensitivity of operational data, which is often subject to strict privacy regulations. Sharing raw data between different stakeholders (e.g., airlines and engine manufacturers) for traditional DA techniques can violate privacy policies and jeopardize proprietary information. Consequently, this prevents the direct application of the DA techniques.

\subsubsection{Interpretability and Model Trustworthiness}
The lack of interpretability in DA models is a significant issue in critical aerospace applications like RUL prediction, where understanding the reasoning behind model predictions is crucial for maintenance decision-making. Black-box models that perform well on paper often lack the transparency needed to be trusted by domain experts.

\subsubsection{Dynamic Changing Operational Environment}
The dynamic operational environment of turbofan engines often results in dynamic domain shifts due to variations in flight routes, weather conditions, and maintenance schedules. This problem limits the generalizability of the adapted model, affecting its performance. Thus, the current DA methods, which relies on target data with static conditions, may struggle to adapt effectively to such evolving conditions.

\subsection{Real-World Case Studies and Industry Adoption}
Several industrial applications highlight the potential of DA techniques for RUL prediction in turbofan engines:
\subsubsection{Cross-Fleet Adaptation for Predictive Maintenance}
\textcolor{black}{A major challenge for airlines operating diverse fleets is maintaining consistent RUL prediction accuracy across different aircraft models and flight conditions.} DA techniques have been used to adapt models trained on one fleet to another, addressing discrepancies in operational patterns. \textcolor{black}{This cross-fleet adaptation reduces the need for collecting large amounts of labeled RUL data from each fleet, thereby lowering operational costs and enhancing maintenance scheduling.}

\subsubsection{Domain Adaptation for New Engine Variants}
When a new variant of an existing engine model is introduced, it may exhibit slightly different degradation patterns due to changes in design or materials. DA methods have been applied to leverage data from previous engine models to predict RUL for the new variant. This adaptation process accelerates the deployment of predictive maintenance models for new engines without extensive data collection and re-training, thus reducing the time to market.

\section{Future Directions}
Building on the challenges discussed in the previous section, we present potential solutions in this section to guide future research directions and advance this field.

\subsection{Privacy-Preserving DA for Turbofan Engine RUL}

To address the mentioned data privacy problem, where operational data from turbofan engines is considered highly sensitive and often subject to stringent data protection regulations, federated Learning (FL) has been proposed as a potential solution by enabling collaborative model training without direct data sharing \cite{cheng2020federated,fink2021artificial,chen2023remaining}. Currently, its integration with deep DA techniques for turbofan RUL prediction remains underexplored. \textcolor{black}{Future research should focus on developing privacy-preserving DA frameworks that leverage FL to transfer temporal degradation knowledge from extensive source domains (e.g., different fleets or engine types) to target domains while maintaining strict data confidentiality. Establishing collaborative platforms involving multiple stakeholders, such as airlines, maintenance providers, and regulatory bodies, could facilitate data sharing and model training in a privacy-compliant manner. Additionally, investigating the use of differential privacy and secure multi-party computation (SMPC) could further enhance data security in FL-based DA frameworks. These efforts would enable shared knowledge across diverse operational contexts, improving the reliability and robustness of RUL predictions for turbofan engines.}

\subsection{Interpretable DA for Critical Aerospace Systems}
As discussed eariler, the lack of interpretability in DA models remains a major obstacle to their adoption in aerospace applications. Thus, current research in Explainable AI (XAI) primarily focusing on visual tasks and simple time-series models \cite{longo2020explainable,youness2023explainable}, can provide effective solution by offering limited utility for complex, multi-sensor datasets used in turbofan engine health monitoring. \textcolor{black}{Future research should develop interpretable DA methods tailored to RUL prediction by incorporating self-explainable architectures or domain-specific feature attribution techniques. Collaborating with domain experts to design visualization tools could elucidate how engine health conditions change across source and target domains. Furthermore, exploring the integration of causal inference methods could help identify the root causes of domain shifts and their impact on RUL predictions. Such models would enhance the transparency of deep learning predictions, increasing trust and adoption in high-stakes aerospace maintenance scenarios.}

\subsection{Few-Shot DA for Evolving Operational Conditions}
To deal with the dynamic operational environment of turbofan engines, few-shot DA methods offer a viable solution. By leveraging a limited set of labeled data from new scenarios, few-shot DA can easily update a model without requiring too many samples \cite{yang2024few}. Currently, the application of few-shot learning in turbofan RUL prediction is still at an early stage, with challenges in generalizing from scarce labeled samples. \textcolor{black}{Future research should focus on meta-learning approaches and adaptive feature representations that can rapidly adapt to new flight patterns or environmental variations using minimal data. Investigating the use of synthetic data generation techniques, such as generative adversarial networks (GANs), could augment few-shot datasets and improve model generalization. Collaborating with industry partners to collect and annotate real-world operational data under diverse conditions would enable the validation of few-shot DA methods in practical settings. This direction is particularly vital for RUL prediction, where acquiring labeled data for each unique operational condition is costly and time-consuming.}

\subsection{Foundation Models for Evolving Operational Conditions}
Foundation models offer a novel approach for addressing evolving operational conditions. Pretrained on large-scale, multi-domain datasets, these models serve as universal feature extractors that capture rich temporal and degradation dynamics across varying engine conditions, providing a robust starting point for DA. Their key advantage lies in cross-domain generalization ability, enabling them to adapt to new target domains with limited labeled data or even in unsupervised scenarios. By leveraging extensive pretraining, these models learn shared temporal patterns and degradation behaviors from diverse operational contexts, making them more resilient to the evolving domain shifts that challenge conventional DA methods.

Recent advancements in time-series foundation models have demonstrated the potential for DA in forecasting tasks \cite{jia2024gpt4mts,morales2024developing,cao2023tempo,zhou2023one}. However, directly applying these models to RUL prediction for turbofan engines introduces unique challenges. Unlike standard time-series forecasting that focuses on predicting future values, RUL prediction involves forecasting the remaining time to failure under varying operational conditions and unpredictable failure mechanisms. \textcolor{black}{To address these challenges, future research should develop foundation models specifically tailored for DA in RUL prediction, incorporating DA layers or meta-learning strategies that can dynamically adapt to new engine conditions using limited data. Collaborating with research institutions and industry leaders to create large-scale, multi-domain datasets for pretraining foundation models could accelerate progress. Additionally, investigating the integration of physics-informed learning could improve the generalization of foundation models across diverse operational environments. By leveraging these approaches, foundation models could facilitate cross-domain RUL adaptation without extensive retraining, making them highly suitable for predictive maintenance across multiple fleets and operational environments.}

\color{black}\section{Conclusion}

This survey presents a focused review of Domain Adaptation (DA) techniques for Remaining Useful Life (RUL) prediction of turbofan engines, a critical component in advancing predictive maintenance strategies in aviation. \textcolor{black}{By introducing a novel taxonomy tailored to the unique characteristics of turbofan engines, we analyze the current state-of-the-art methods and offer a comprehensive understanding of existing work in this field from multiple perspectives.} Our analysis not only highlights the potential of DA to address the limitations of conventional data-driven approaches but also identifies future research directions to enhance aviation safety and reliability. Additionally, we present an open-source collection of DA techniques specifically for turbofan engines, which contributes to the practical application and testing of these methods, fostering innovation and collaboration within the research community. By bridging theoretical exploration with practical implementation, this survey aims to stimulate advancements in the field, ultimately leading to more efficient, cost-effective, and reliable operations for turbofan engines.

% \section*{Acknowledgments}
% This should be a simple paragraph before the References to thank those individuals and institutions who have supported your work on this article.

\bibliographystyle{IEEEtran}
\bibliography{ref}

\end{document}